\documentclass[10pt,twocolumn,letterpaper]{article}

\usepackage{iccv}
\usepackage{times}
\usepackage{epsfig}
\usepackage{graphicx}
\usepackage{amsmath}
\usepackage{amssymb}
\usepackage{multirow}
\usepackage{subfigure}
\usepackage{makecell}
\usepackage{mathrsfs}

\usepackage[pagebackref=true,breaklinks=true,letterpaper=true,colorlinks,bookmarks=false]{hyperref}

\iccvfinalcopy 

\begin{document}

\title{ELSD: Efficient Line Segment Detector and Descriptor}

\author{Haotian Zhang\\
Megvii Technology\\
{\tt\small zhanghaotian@megvii.com}
\and
Yicheng Luo\\
Megvii Technology\\
{\tt\small luoyicheng@megvii.com}
\and
Fangbo Qin\\
Institute of Automation, CAS\\
{\tt\small qinfangbo2013@ia.ac.cn}
\and
Yijia He\\
{\tt\small heyijia2016@gmail.com}
\and
Xiao Liu\\
Megvii Technology\\
{\tt\small liuxiao@megvii.com}
}

\maketitle
\ificcvfinal\thispagestyle{empty}\fi

\begin{abstract}
   We present the novel Efficient Line Segment Detector and Descriptor (ELSD) to simultaneously detect line segments and extract their descriptors in an image. Unlike the traditional pipelines that conduct detection and description separately, ELSD utilizes a shared feature extractor for both detection and description, to provide the essential line features to the higher-level tasks like SLAM and image matching in real time. First, we design the one-stage compact model, and propose to use the mid-point, angle and length as the minimal representation of line segment, which also guarantees the center-symmetry. The non-centerness suppression is proposed to filter out the fragmented line segments caused by lines' intersections. The fine offset prediction is designed to refine the mid-point localization. Second, the line descriptor branch is integrated with the detector branch, and the two branches are jointly trained in an end-to-end manner. In the experiments, the proposed ELSD achieves the state-of-the-art performance on the Wireframe dataset and YorkUrban dataset, in both accuracy and efficiency. The line description ability of ELSD also outperforms the previous works on the line matching task.
\end{abstract}

\section{Introduction}

\begin{figure}[t]
\begin{center}
   \includegraphics[width=0.9\linewidth]{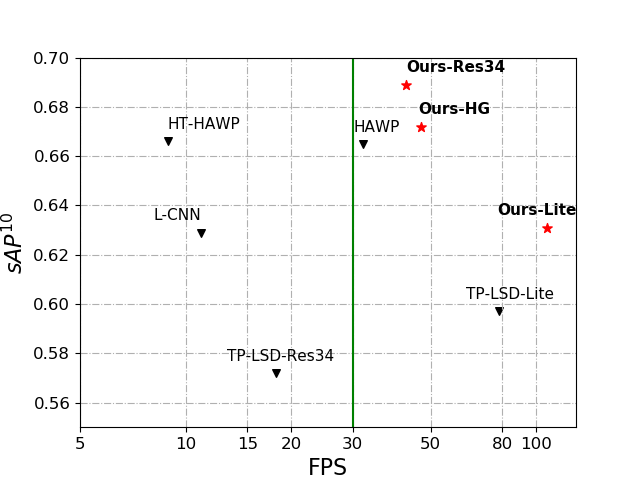}
\end{center}
   \caption{Inference speed (FPS) and accuracy ($sAP^{10}$) on Wireframe dataset.}
\label{FPS}
\end{figure}

Image representation is an essential issue for many computer vision tasks such as SLAM,  Structure-from-Motion (SfM), and image matching. Local point features \cite{superpoint,SIFT,orb} are widely used in these tasks, and recently the researchers have been exploring the usage of structural features for the better geometric representation\cite{plvio,planar_SLAM,RGB-D_SLAM_for_planar,Line_Correspondences,Line-Based_Map}. Line segments are the most widely seen structural features in man-made environments. The reliable extraction of line segments and the matching across frames are important for the aforementioned tasks.

Recently, the convolutional neural networks (CNN) based line segment detection models have significantly outperformed the traditional methods. The models\cite{HAWP,PPGNET,LCNN} consist of two stages.
They first detect junctions and then generate line segment proposals and finally feed the embedding of each line segment into a classifier. Although these two-stage methods can achieve high performance, their running speed cannot satisfy real-time applications. TP-LSD\cite{TP-LSD} first realizes the compact one-stage detection by introducing the Tri-points representation of line segment. However, TP-LSD predicts the two end-points separately and does not leverage the center-symmetric characteristics of the line segment. Thus, the predicted root-point might not be the mid-point of the two predicted end-points, and even the three points might be not co-linear. Moreover, the prediction of the root point is ambiguous especially when the lines intersect with each other so that many false root-points belonging to the fragmented line segments are detected.  Besides, TP-LSD does not differentiate hard and easy examples during training. Some hard root points of line segments may not be properly detected.  

Line segment descriptor is required to represent the line segment in a high-dimensional metric space, and the same line in two adjacent frames should be close in this metric space. There exist some CNN-based line descriptors\cite{WLD,DLD,LLD}. However, these line descriptors are designed individually, and not yet tightly coupled with the line segment detector. It is also time-consuming to execute detection and description separately.

To this end, we propose ELSD that simultaneously predicts line segments and inferences line descriptors in an end-to-end fashion. 1) We introduce the one-stage architecture that utilizes the Center-Angle-Length (CAL) representation to vectorize a line segment. Our line detector consists of two module: $(i)$ localization module and $(ii)$ regression module. 2) Since the mid-points might be ambiguous for detection when lines intersect, as shown in Figure \ref{centerness}b, we introduce the line-centerness to filter the false mid-points belonging to fragmented line segments and adopt modified focal loss\cite{Focal} to focus more on the mid-points of hard cases. 3) In the regression module, the geometric maps are predicted to provide the rotation angles and lengths. Moreover, we refine the position of the midpoints by predicting the fine offsets to compensate for the localization accuracy. 4) In the line descriptor branch, we obtain the descriptor of each predicted line segment by line pooling. The descriptor is trained learned by random homography-based self-supervision. The pipeline of ELSD is shown in Figure \ref{framework}.


In summary, the main contributions are as follows:

\begin{itemize}
\item We present a pipeline that simultaneously detects line segments and inferences line descriptors in an end-to-end fashion. To the best of our knowledge, this is the first work that unifies line detector and descriptor in a compact neural network. The major computation in the backbone is shared by the two tasks, and the two task branches can be jointly training, with negligible loss on detection performance.

\item We utilize the Center-Angle-Length (CAL) representation to encode a line segment that has only four parameters to predict. To overcome the detection ambiguity when lines intersect, we proposed the non-centerness suppression mechanism to remove the mid-points of fragmented line segments. The midpoint position is further refined by using the offset regression so that the line segment localization is more precise.

\item Our proposed ELSD obtains state-of-the-art performance in both accuracy and efficiency on the Wireframe and YorkUrban datasets. Moreover, the light version of our model achieves the speed of 107.5 FPS on a single GPU (RTX2080Ti) with comparable performance.


\end{itemize}

\section{Related Works}
\subsection{Line Segment Detection}
Deep learning-based line segment detection methods has attracted great attention due to the remarkable performances\cite{wireframe_cvpr18,TP-LSD,HAWP,LCNN,3Dwireframe}. 
AFM\cite{AFM} presented regional partition maps and attraction field maps of line segment maps, followed by a squeeze module to generate line segments. L-CNN\cite{LCNN} first proposed a two-stage pipeline for wireframe parser. It predicts junction map to generate line proposals and utilize the LoI-pooling to gather feature of the proposals. Then a line verification network classifies proposals and removes false lines. PPGNet\cite{PPGNET} used a graph formulation to represent the relation between junctions. HAWP\cite{HAWP} proposed a 4-D holistic attraction field map for generating line proposals and refine the proposals with junction heat maps. HT-HAWP\cite{HT-HAWP} combined Hough transform and HAWP model, obtaining excellent results in line segment detection. As the first one-stage line segment detector, TP-LSD\cite{TP-LSD} proposed a Tri-Points representation to encode line segments and predicted two endpoints of each line segment in an end-to-end manner. LETR\cite{LETR} applied transformers for line segment detection from coarse-to-fine grained. Our ELSD has a similar pipeline with TP-LSD. We encode a line segment by CAL representation, and can directly detect possible semantic line segments in the image without additional classification.

\subsection{Object Detection}
The recent surge of some keypoint-based object detectors has achieved remarkable performance. CornerNet\cite{CornerNet} formulated each object by a pair of corner keypoints and grouped all the detected corner keypoints to form the final detected bounding box, which requires more complicated post-processing.
CenterNet\cite{Centernet} models an object by the center point of its bounding box, and uses keypoint estimation method to find center points and regresses to its size. FCOS\cite{FCOS}  treats all the pixels with an object as candidate position and proposed center-ness to represent the importance of all the candidate positions. PolarNet\cite{polarnet} learns corner pairs based on polar coordinates and avoids the large variance of learned offsets in Cartesian coordinate. Such keypoint-based methods have good detection capabilities with a  fast speed and brief structure. Motivated by these, we proposed a new line segment representation and further designed a keypoint-based line segment detector.  

\begin{figure*}[t]
	\begin{center}
		\scalebox{0.8}{
			\includegraphics[width=0.9\linewidth]{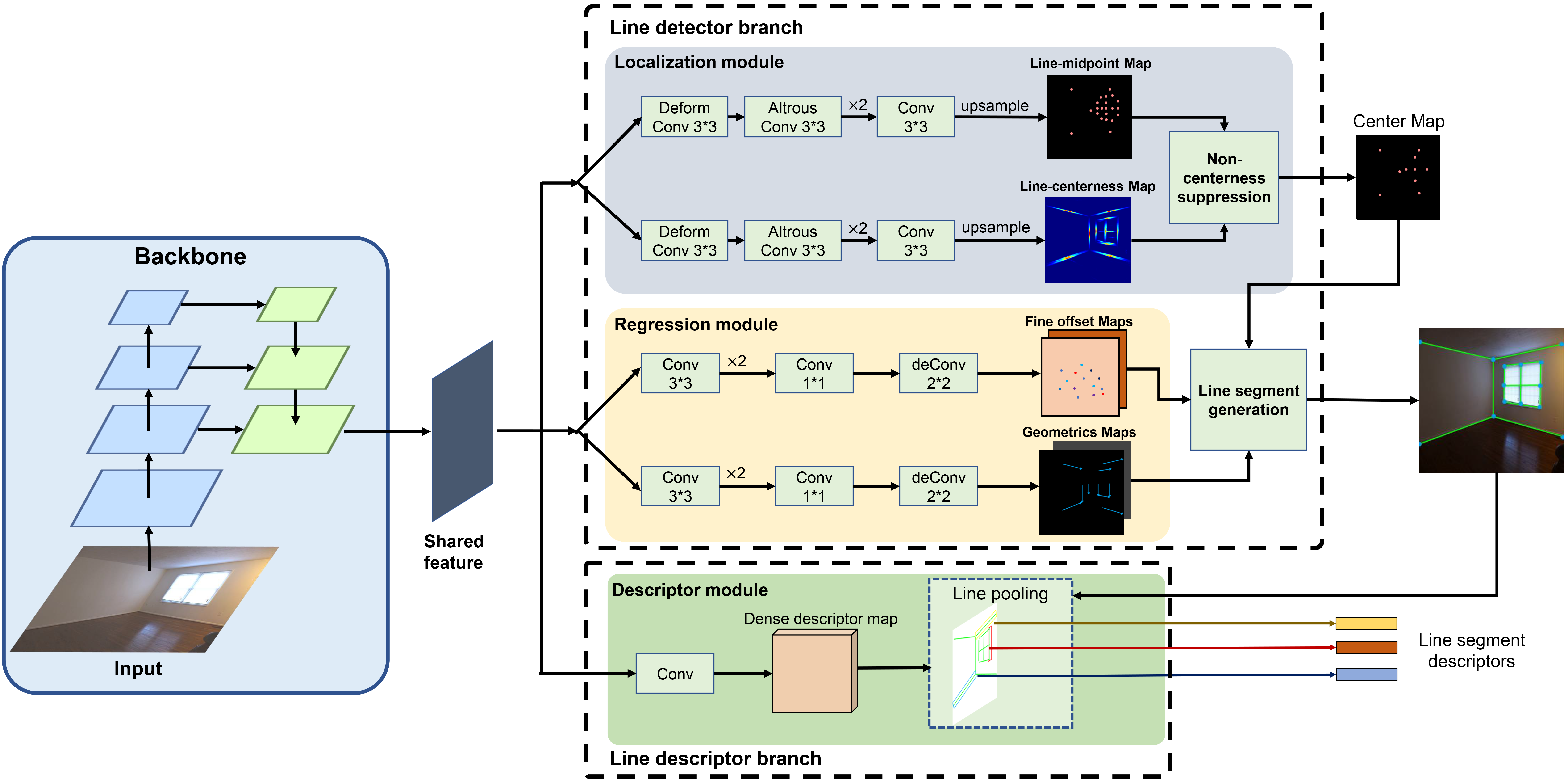}
		}
	\end{center}
	\caption{Illustration of the architecture of our proposed ELSD. It consists of three components: backbone, line detector branch and line descriptor branch. See text for details.}
	\label{framework}
\end{figure*}

\subsection{Line Description}
Like descriptor-based keypoint matching\cite{superpoint,SIFT,orb}, line matching is also based on comparing the descriptors of the same line segments in two frames. MSLD\cite{MSLD} constructs the line descriptors by counting the mean and variance of the gradients of pixels in the neighbor region of a line segment. LBD\cite{LBD} proposes a line-band descriptor that computes gradient histograms over bands with more robustness and efficiency. Recently, some deep learning-based methods are used in learning line descriptors. LLD\cite{LLD} and DLD\cite{DLD} use the convolution neural network to learn the line descriptors and achieve remarkable performance. 

\section{Methods}
\subsection{Line Representation}
Line segments have two characteristics: 1) Due to the center-symmetry, the mid-point determines the location of the line segment, then the geometric feature is determined by the angle and length. 2) Since a line segment is straight, its direction can be consistently measured from a local part of it, which is easier to learn and requires a small receptive field. Therefore, we propose the Center-Angle-Length (CAL) representation to vectorize a line segment, which only has four parameters: 2D coordinates, rotation angle, and total length. In comparison, the Tri-points representation in TP-LSD\cite{TP-LSD} has six parameters to predict, which is redundant, and the prediction results might not satisfy the center-symmetry.


With angle $ \theta $, length $ \rho $, and center point $\left[\begin{array}{l}x_{c} \\ y_{c}\end{array}\right]$, the two endpoints of the line segment are given by,

\begin{small}
\begin{flalign}
\label{polar_eq}
\left[\begin{array}{l}x_{s} \\y_{s} \end{array}\right]=\left[\begin{array}{cc}
x_{c}  \\y_{c}\end{array}\right] + \frac{\rho}{2}\left[\begin{array}{c}cos\ \theta \\sin\ \theta
\end{array}\right] \nonumber\\
\left[\begin{array}{l}x_{e} \\y_{e} \end{array}\right]=\left[\begin{array}{cc} x_{c}  \\
y_{c}\end{array}\right] - \frac{\rho}{2}\left[\begin{array}{c}cos\ \theta \\sin\ \theta\end{array}\right] \end{flalign}.
\end{small}

\subsection{Overall Network Architecture}
As shown in Figure \ref{framework}, our proposed ELSD consists of a backbone, a line detector branch, and a line descriptor branch. Our backbone is a U-shape network that consists of an encoder and two decoder blocks. The backbone takes an image of size $ 3\times 512\times 512 $ as input and outputs the shared feature with a size of $ 128 \times 128\times 128 $. After the backbone, the architecture splits into two parts: one for line detector and the other for line descriptor. The line detector branch can predict line segments from an image. We can further obtain line descriptors by feeding both shared feature and predicted line segments into the line descriptor branch. ELSD can produce line segments and further extract fixed dimensional descriptors of the line segments in a single forward pass. Moreover, unlike the traditional pipeline that first detects line segments, then computes line descriptors, ELSD shares most of the parameters between these two tasks, which reduces the computation cost and improves the compactness. 


\subsection{Line Detector Branch}
Our line detector branch takes the shared feature from the backbone as input and splits into two modules: 1) Localization module, which consists of a line-midpoint detection head and a line-centerness detection head. In Non-Centerness-Suppression (NCS), the two heads are combined to get a more accurate center detection; 2) Regression module, which contains a geometrics regression head and a fine offset regression head. The outputs of the regression module are a pair of geometrics maps that consists of $ (\rho,\theta) $ and a pair of fine offset maps. Finally, the outputs of two modules are combined together to generate the mid-points with two symmetrical endpoints as the line segment detection results.

\subsubsection{Localization Module}
Similar to TP-LSD\cite{TP-LSD}, we use a deformable convolution, two atrous convolution (dilation rate=2) and a standard convolution layers to obtain the adaptive spatial sampling and a large receptive field, to predict the mid-point map. Furthermore, we leverage the line-centerness, i.e. how close an on-line point lies to the mid-point, to distinct the mid-points of the entire lines and the fragmented lines. The line-centerness is calculated by,
\begin{small}
\begin{equation}
    P_{centerness} = \sqrt{\frac{min(d_1,d_2)}{max(d_1,d_2)}}
    \label{centerness_eq}
\end{equation}
\end{small}
where $ d_1,d_2 $ are the distances from a point on the line segment to the two end-points, respectively. Apparently, $P_{centerness}$ equals 1 when the point is midpoint and decreases to 0 when the point approximates the end-points.  
\begin{figure}[t]
\begin{center}
   \includegraphics[width=0.9\linewidth]{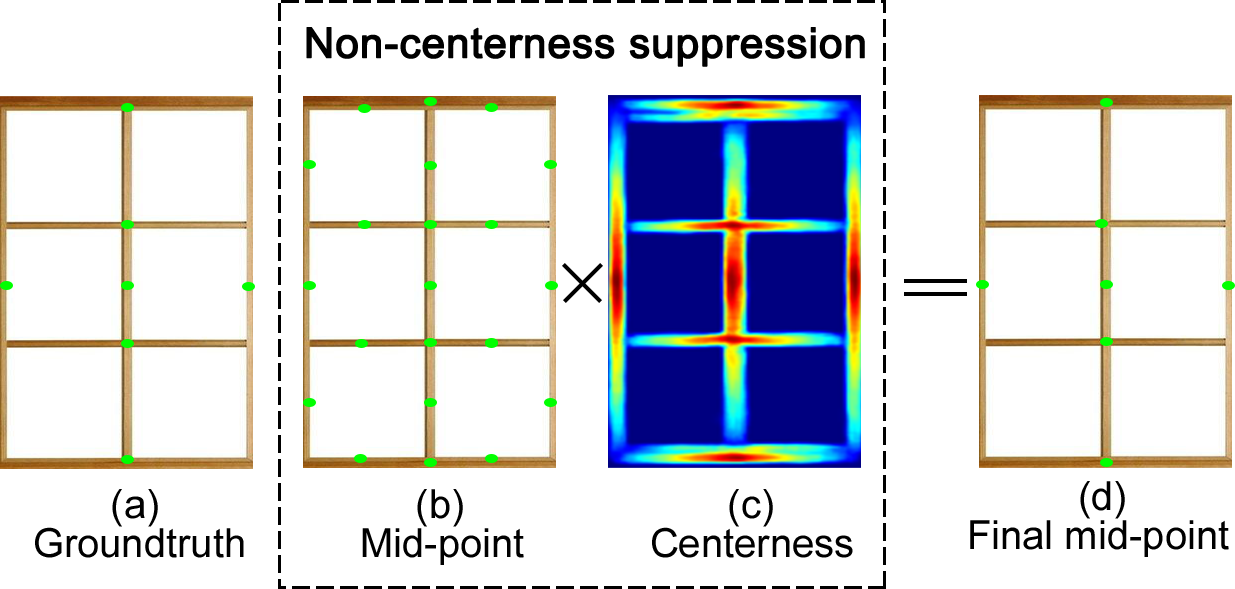}
\end{center}
   \caption{Illustration of Non-Centerness-Suppression (NCS). (b) and (c) show the predicted mid-point map and centerness map, respectively.}
\label{centerness}
\end{figure}

The line-centerness module has the same architecture as the localization module. 
Denote the predicted line-midpoint map and line-centerness map as $ \widehat{P}_{mid} $ and $ \widehat{P}_{centerness} $, respectively. As shown in Figure \ref{centerness}, we propose the Non-Centerness Suppression (NCS) to filter false local midpoints belonging to fragmented line segments, and obtain a more accurate center confidence map $ \widehat{P} $, as given by,

\begin{equation}\label{center_eq}
\small
    \widehat{P} = \widehat{P}_{mid} \times \widehat{P}_{centerness}^{0.5}
\end{equation}

The effectiveness of NCS is explained as follows. The midpoint detection is to obtain the exact positions but is prone to false detection caused by lines intersection. As shown in Figure \ref{centerness}, when a line segment is intersected with another line, its two endpoints and the intersection point form two shorter fragmented line segments. Although the mid-points of these fragmented line segments are not annotated as ground truths and are not expected to be detected, the detector tends to detect them because the fragmented line segments satisfy the definition of line segment. Differently, as visualized in Figure \ref{centerness}, line-centerness is not exact but provides a non-local distribution along the global line segment. The non-local distribution is more significant to inference and contains the global structure information of the potentially intersected lines. Namely, the midpoints can only mark a line segment without the awareness of the global structure, and the line-centerness map can further encode the global structure information with a non-local non-linear multi-peak 2D distribution. Therefore, the line mid-point map and line-centerness map are fused by Eq \ref{center_eq} to suppress the false detection and get the final mid-points. Thus the ambiguity problem met by TP-LSD is effectively alleviated.

\subsubsection{Regression Module} \label{polar_sec}
Our regression module consists of two heads: a fine offset regression head and a geometrics regression head. The fine offset regression head is used to predict the offset of the center caused by the downsampling ratio. The refined sub-pixel mid-point can be obtained by just add the corresponding offset to the position of the predicted mid-point. The geometrics regression head can predict angle and length with respect to the midpoint. Both of our regression heads contain two $ 3\times3 $, a $ 1\times1 $ convolutional layers, and a deconvolutional layer. The deconvolutional layer is used to restore the size of the output map to $ 256\times256$. We can index the related angle $ \theta $ and length $ \rho $ by the center position $ (x_c,y_c) $ on the output map. Then a line segment can be obtained by Eq \ref{polar_eq}.

We utilize the CAL representation rather than Cartesian coordinates representation because the angle belongs to the geometric attributes of the line segment itself. Since the angle information can be perceived from a local part of the line segment, it is easier and more precise to predict the angle than the coordinates. We have done experiments to compare CAL representation and Cartesian coordinate representation under the same settings in Section \ref{ablation_sec}.

\begin{figure}[t]
	\begin{center}
		\scalebox{1.1}{
			\includegraphics[width=0.9\linewidth]{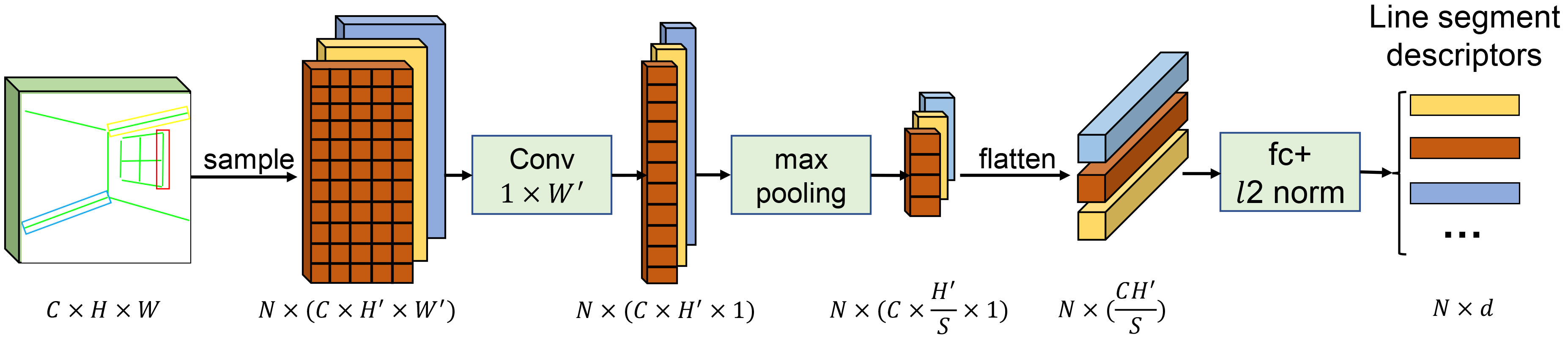}
		}
	\end{center}
	\caption{Illustration of Line Pooling. See text for more details.}
	\label{linepooling}
\end{figure}
\subsection{Line Descriptor Branch}
Given a set of line segments, the purpose of the line descriptor branch is to learn a fixed-length descriptor for each line segment, which is used to distinguish different line segments according to the distance between their descriptors. We first apply two $3\times 3$  stride-1 convolution on the shared feature map from backbone. Then this intermediate feature map is resized to $256 \times 256$ by bilinear interpolation. The resulting feature map named dense descriptor map is used in the following Line Pooling.


\textbf{Line Pooling:} Similar to RoIPool\cite{roipool} and RoIAlign\cite{roialign} used in object detection, the Line Pooling is used to squeeze the rotated narrow ROI to a descriptor vector. As shown in Figure \ref{linepooling}, the RoI of a line segment is defined as a rotated bounding box centered at the line segment, with the same length and angle as the line segment. The width of the RoI is a hyperparameter that depends on the desired size of the receptive field. Then we crop a fixed-size line feature map by sampling from the dense descriptor map using bilinear interpolation. Assuming there exist $N$ candidate line segments and each line feature map has the size of $C\times H'\times W'$, in which $C$ is the channel dimension of the dense descriptor map and $H', W'$ represent the height and width of the line feature map respectively. We further apply a $1\times W'$ stride-1 depth-wise convolution as well as a stride-$S$ max pooling to the line feature map. Finally, the resulting feature vector is flattened and fed into a fully connected layer and then normalized, producing the final descriptor with a fixed length denoted as $d$.

\begin{figure}[]
\begin{center}
\centering
\includegraphics[width=\linewidth]{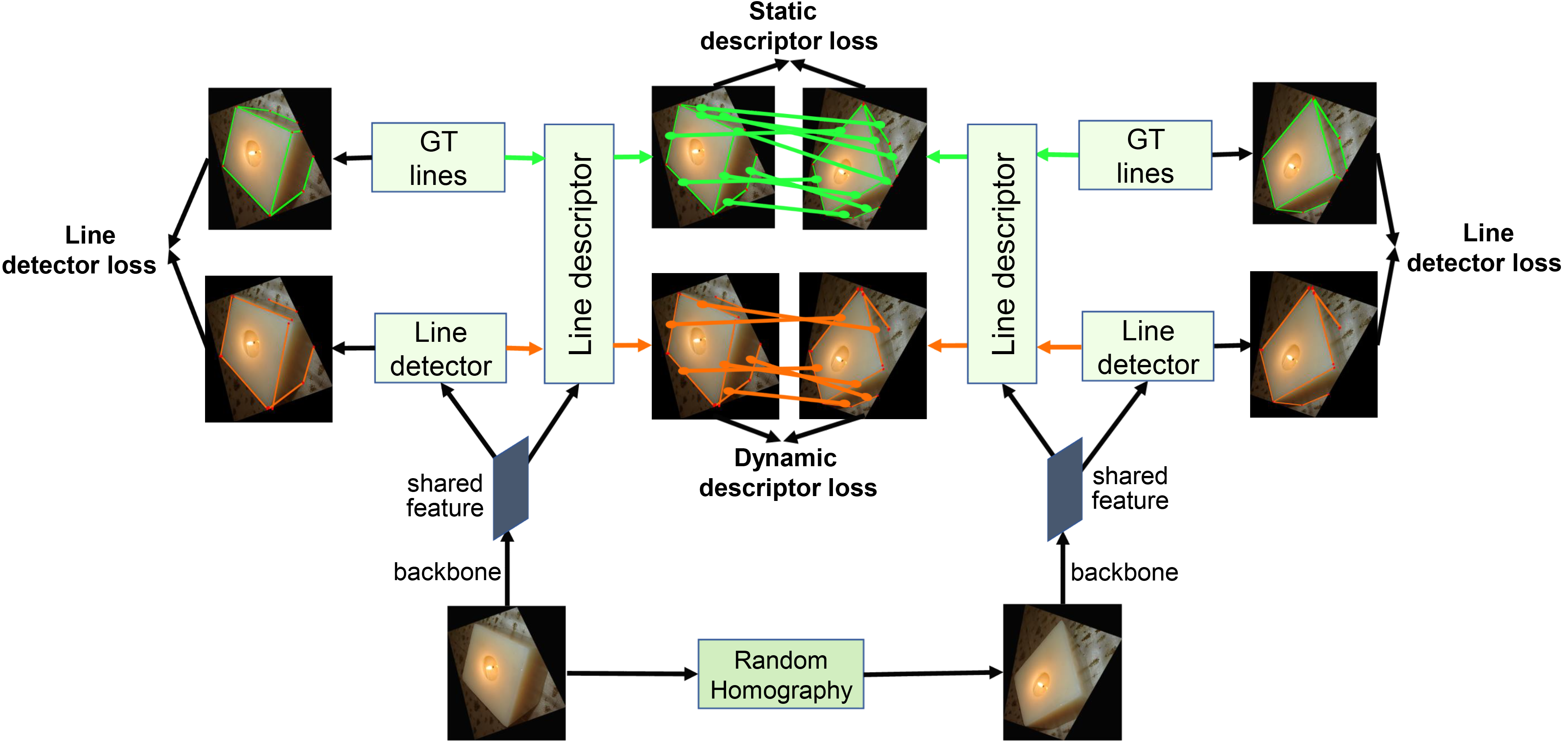}
\end{center}
 \caption{Training framework. The random homography is used to realize self-supervised training. See text for details.}
    \label{training_fig}
\end{figure}

\textbf{Self-supervised learning:} Similar to \cite{superpoint}, we apply random homographies on an image to produce a paired image with different views of the same scene, assuming that planar scenes or distant scenes are common in the real environment. The homography transformation that we used is composed of a set of transformations such as translation, scale, rotation, and perspective distortion, covering most of the viewpoint change caused by camera motion. After applying random homography on the input image, we can obtain the exact image-to-image transformation. So we can label matched or unmatched line segments just by transforming the endpoint of the line segment from one image to another and checking whether the distance of two corresponding endpoints is close enough.\label{static_and_dynamic}

When training from scratch, inspired by the Line Sampling Module of L-CNN\cite{LCNN} that adopts static line sampler and dynamic line sampler to train the classifier, we use static line segments and dynamic line segments to train the descriptors. In the training stage, the static line segments are the annotated ground-truths, and the dynamic line segments are those predicted by the detection branch which changed as the model training proceeds. Because the line segment detection are not confident at the early training stage, we only use the detected line segments that are close enough to ground-truths as the dynamic line segments. 
Note that for training line descriptor branch, the proposed ELSD is trained on mini-batches of image pairs. We can obtain the ground truth correspondence of a pair of image's static line segments set during data preparation. The ground truth correspondence of a pair of dynamic line segments can be given by its closest static line segments. If the closest static line segments of a pair of dynamic line segments are matched, we then label this dynamic pair as a match and otherwise non-match. 
The whole training process of ELSD is shown in Figure \ref{training_fig}. 
To sum up, the training with the static line segments helps to cold-start the training of descriptors at the beginning. The training with dynamic line segments helps to couple the descriptor with the actual prediction of the detector.

\subsection{Loss Functions}
\subsubsection{Total Loss}
The total loss to train ELSD is composed of the line detector loss $\mathcal{L}_{p}$ and line descriptor loss $\mathcal{L}_{d}$. Note that the input of ELSD is a pair of images with random homographies, which have both ground truth line segments, as well as the ground truth correspondence of ground truth line segments and predicted line segments. This allows us to optimize the two losses simultaneously. Given a pair of image, $(I^A,I^B)$, and the total loss can be represented as: 
\begin{equation}
\mathcal{L}(A,B)=\lambda_{p}(\mathcal{L}_{p}(A)+\mathcal{L}_{p}(B))
+\lambda_{d}\mathcal{L}_{d}(A,B)
\end{equation}
We empirically set $\lambda_{p}=0.9, \lambda_{d}=0.1$ in this work.
\subsubsection{Line Detector Loss}
In the training stage of line detector branch, the outputs of four heads include line midpoint map, line-centerness map, geometrics maps, and fine offset maps. The ground truth of these maps is generated from the raw line segments label. The total loss of line segment detection is shown in Eq \eqref{AllLoss}
\begin{equation}
\begin{split}
    \mathcal{L}_{p}(A)=&\lambda_{mid}\mathcal{L}_{mid}(A)+\lambda_{cen}\mathcal{L}_{cen}(A)+
    \\ &\lambda_{geo}\mathcal{L}_{geo}(A)+\lambda_{off}\mathcal{L}_{off}(A)     
\end{split}
\label{AllLoss}
\end{equation}
where weights $\lambda_{mid,cen,geo,off}=\{25,10,1,3\}$

\textbf{Localization loss:} Given an image $I^A$, for each ground truth midpoint $ p $ with continuous value,
we construct the midpoint confidence map $ P\in[0,1]^{H\times W \times 1} $ with four pixels near the midpoint by flooring and ceiling and we denote the selected pixels set by $ v $. The 2D Gaussian kernel $ G_{xy}=$exp$ (-\frac{(x-p_x)^2+(y-p_y)^2}{2\sigma^2}$) is then used to compute each confidence of the pixels in $ v $. Then we normalize these confidence by dividing the max value of $v$. If the confidence of a pixel is assigned more than one time, we keep the max value of it. The overall process is described by,
\begin{equation}
P_{xy} = max(\frac{G_{xy}}{\underset{(i,j)\in v}{max}G_{ij}},P_{xy})
\end{equation}
Then we followed CornerNet\cite{CornerNet} to use a variant of focal loss:
\begin{equation}
\footnotesize
    \mathcal{L}_{mid}(A) = \frac{-1}{N}\sum_{xy}^{HW}
    \begin{cases}
    (1-\widehat{P}_{xy})^\alpha \mbox{log}(\widehat{P}_{xy}), & \mbox{if } P_{xy}=1 \\
    \begin{split}
     (1&-P_{xy})^\beta(\widehat{P}_{xy})^\alpha \\
    & \mbox{log}(1-\widehat{Y}_{xy}), 
    \end{split}
    & \mbox{otherwise}
    \end{cases}
\end{equation}
where $ \alpha $ and $ \beta $ are hyper-parameters and $N$ is the number of midpoints in an image. We set $ \alpha=2 $ and $ \beta=4 $.

 According to Eq \eqref{centerness_eq}, we can obtain ground truth centerness map. Then we use weighted Binary Cross Entropy (BCE) loss denoted as $\mathcal{L}_{cen}$ to supervise the learning process of the centerness.

\textbf{Regression loss:} Suppose the ground truth angle, length is $(\theta,\rho)$ and the corresponding predicted angle, length is $(\widehat{\theta},\widehat{\rho})$. We use $L1$ loss and smooth $L1$ loss as the geometrics regression loss which is defined as
\begin{equation}
 \mathcal{L}_{geo}(A) = \lambda_{ang}L1(\theta,\widehat{\theta})+\lambda_{len}SmoothL1(\rho,\widehat{\rho})
\end{equation}

where $\lambda_{ang,len}=\{300,10\}$. Besides, to recover the discretization error of midpoint coordinate caused by downsampling with ratio $ s$, we additionally predict the fine offset maps $ \widehat{O} $ for each midpoint. The offset is trained with $\mathscr{l1}$ loss.
\begin{small} 
\begin{equation}
    \mathcal{L}_{off}(A) = \frac{1}{N}\sum_{k=1}^N|\widehat{O}-(p-\left
    \lfloor \frac{p}{s} \right\rfloor)|
\end{equation}
\end{small}
Note that only the midpoints where the confidence score of the ground truth equals 1 are involved in regression loss calculation.

\begin{table*}[t]
\resizebox{\textwidth}{32mm}{
\begin{tabular}{|l|c|c|ccc|cc|ccc|cc|c|}
\hline
\multicolumn{1}{|c|}{\multirow{2}{*}{Method}} & \multicolumn{1}{|c|}{\multirow{2}{*}{ \makecell[l]{Input \\size}    }} & \multirow{2}{*}{Backbone} & \multicolumn{5}{c|}{Wireframe}     & \multicolumn{5}{c|}{YorkUrban}     & \multirow{2}{*}{FPS} \\ \cline{4-13}
\multicolumn{1}{|c|}{}       &\multicolumn{1}{|c|}{}     &                           & sAP$^5$ & sAP$^{10}$ & sAP$^{15}$ & AP$^H$  & F$^H$   & sAP$^5$ & sAP$^{10}$ & sAP$^{15}$ & AP$^H$  & F$^H$   &                      \\ \hline
LSD\cite{LSD}             & 320                             & /                         & /    & /     & /     & 55.2 & 62.5 & /    & /     & /     & 50.9 & 60.1 & 100                \\ \hline
AFM\cite{AFM}             & 320                              & U-Net                     & 18.5 & 24.4  & 27.5  & 69.2 & 77.2 & 7.3  & 9.4   & 11.1  & 48.2 & 63.3 & 12.8                 \\ \hline
DWP\cite{wireframe_cvpr18}             & 512                         & Hourglass                 & 3.7  & 5.1   & 5.9   & 67.8 & 72.2 & 1.5  & 2.1   & 2.6   & 51   & 61.6 & 2.2                 \\ \hline
LETR\cite{LETR}            & 512                         & ResNet101                 & /    & 65.2  & 67.7  & 86.3 & \textbf{83.3} & /    & 29.4  & 31.7  & 62.7 & 66.9 & /                    \\ \hline
TP-LSD-Lite\cite{TP-LSD}     & 320                         & ResNet34                  & 56.4 & 59.7  & /     & /    & 80.4 & 24.8 & 26.8  & /     & /    & \textbf{68.1} & 78.2                 \\ \hline
TP-LSD\cite{TP-LSD}          & 512                         & ResNet34                  & 57.6 & 57.2  & /     & /    & 80.6 & 27.6 & 27.7  & /     & /    & 67.2 & 18.1                 \\ \hline
L-CNN\cite{LCNN}           & 512                         & Hourglass                 & 58.9 & 62.9  & 64.9  & \makecell[l]{80.3 \\82.8$ ^\dagger $} & \makecell[l]{76.9 \\81.3$ ^\dagger $} & 24.3 & 26.4  & 27.5  &  \makecell[l]{58.5 \\59.6$ ^\dagger $} & \makecell[l]{63.8 \\65.3$ ^\dagger $} & 11.1                 \\ \hline
HT-HAWP\cite{HT-HAWP}         & 512                         & Hourglass                 & 62.9 & 66.6  & /     & /    & /    & 25   & 27.4  & /     & /    & /    & 8.9                    \\ \hline 
HAWP\cite{HAWP}            & 512                         & Hourglass                 & 62.5 & 66.5  & 68.2  & \makecell[l]{84.5 \\86.1$ ^\dagger $} & \makecell[l]{80.3 \\83.1$ ^\dagger $} & 26.1 & 28.5  & 29.7  & \makecell[l]{60.6 \\61.2$ ^\dagger $} & \makecell[l]{64.8 \\66.3$ ^\dagger $} & 32.1                 \\ \hline \hline
Ours-Lite         & 256                         & ResNet34                 & 57.4 & 63.1  & 65.5  & 85.6 & 80.2 &  24.3 & 27.4  & 29.3  & \textbf{63.2} & 63.3 & \textbf{107.5}                    \\ \hline
Ours-HG         & 512                         & Hourglass                 & 62.7 & 67.2  & 69.0  & 84.7 & 80.3 &  23.9 & 26.3  & 27.9  & 57.8 & 62.1 & 47                    \\ \hline
Ours-Res34      & 512                         & ResNet34                  & \textbf{64.3} & \textbf{68.9}  & \textbf{70.9}  & \makecell[l]{\textbf{87.2} \\87.3$ ^\dagger $}  & \makecell[l]{82.3 \\83.1$ ^\dagger $} & \textbf{27.6} & \textbf{30.2}  & \textbf{31.8}  & \makecell[l]{62.0 \\62.6$ ^\dagger $} & \makecell[l]{63.6 \\64.8$ ^\dagger $}     & 42.6                  \\ \hline
\end{tabular}
}
\caption{Comparison experiments on line segment detection. '/' means the values are not reported in the related paper. '$^\dagger$' means the post-processing scheme proposed in L-CNN\cite{LCNN} is used.} 
\label{CompareWithLSD}
\end{table*}

\begin{figure*}[htbp]
\centering
\includegraphics[width=0.24\textwidth]{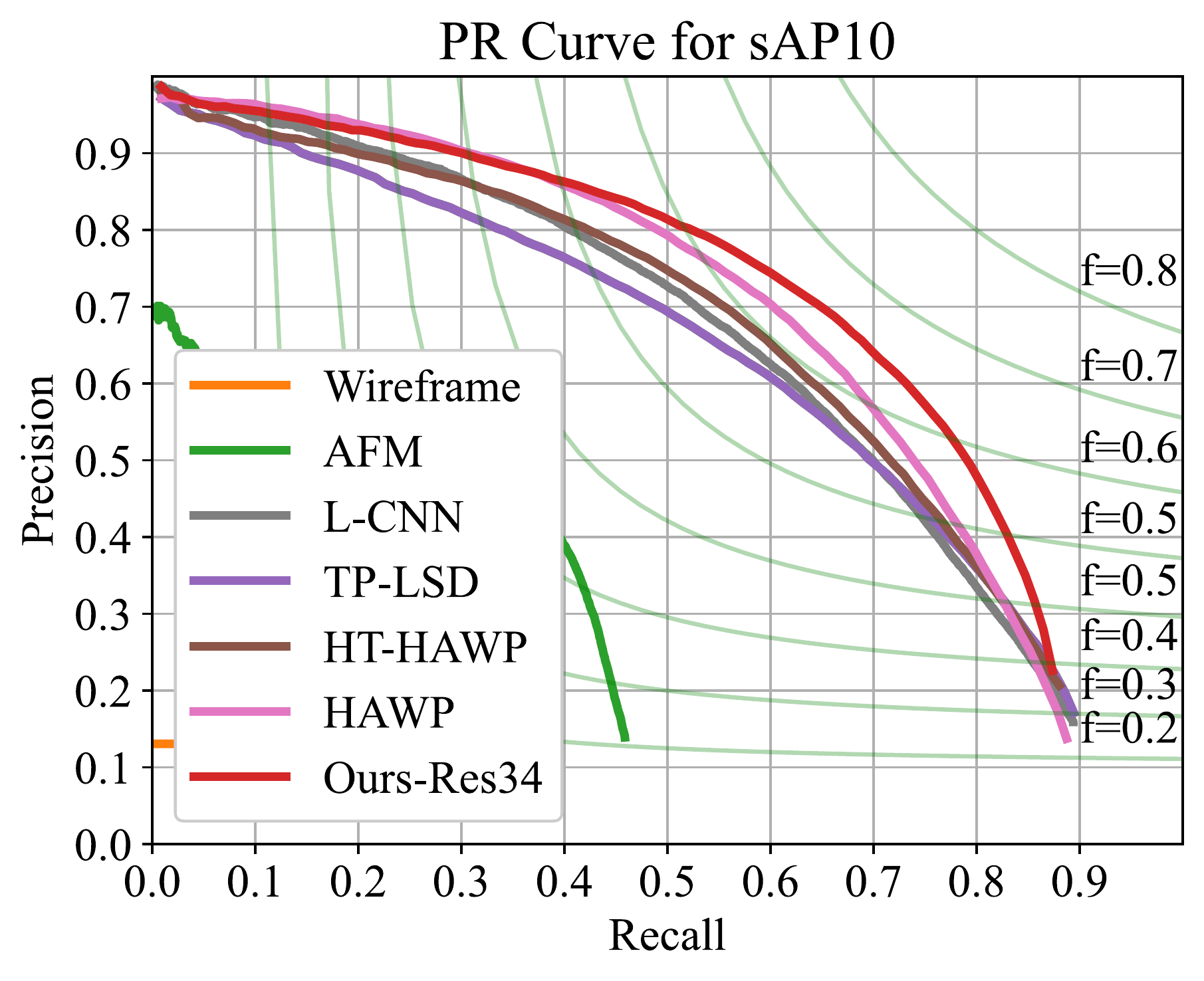}
\includegraphics[width=0.24\textwidth]{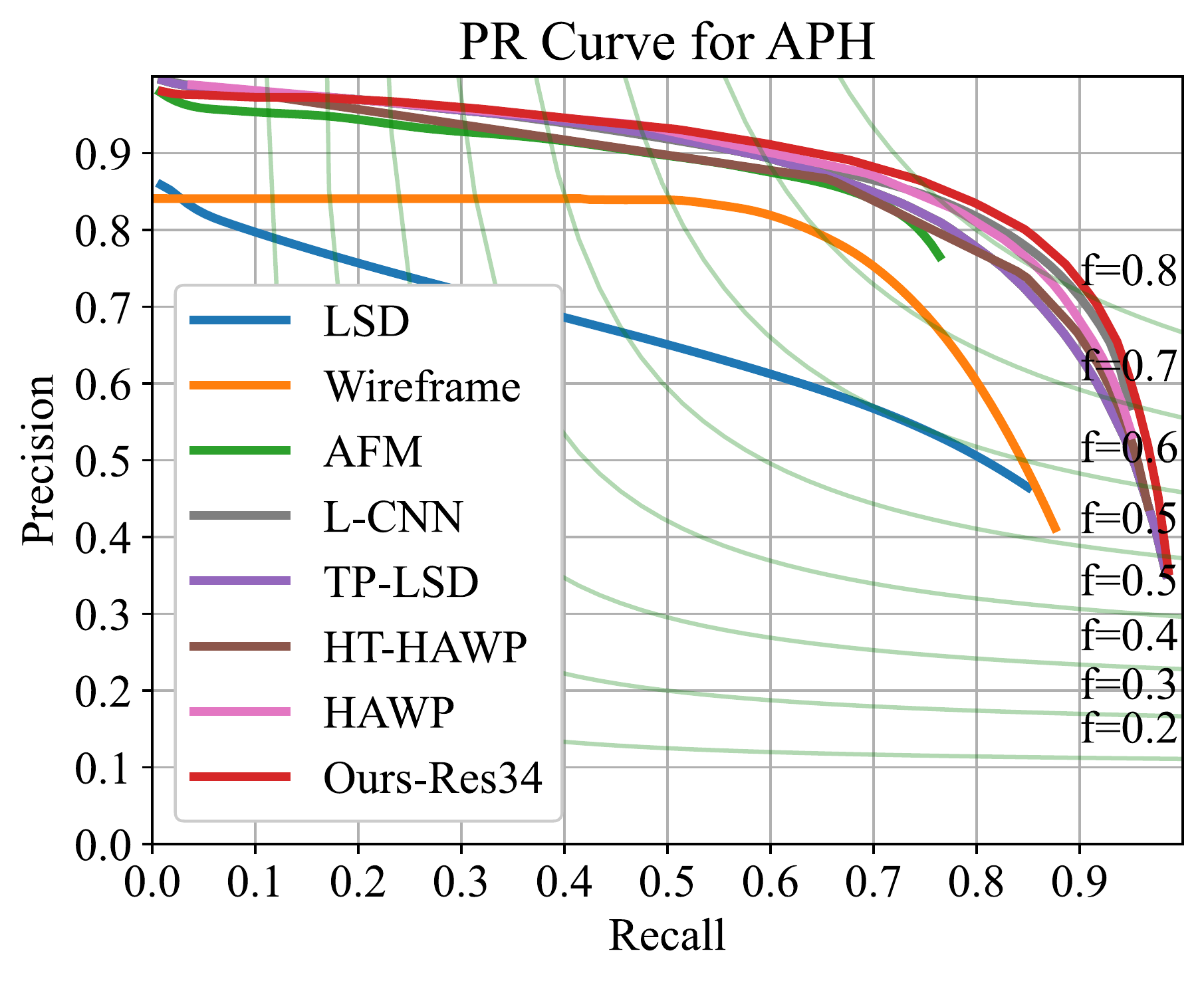}
\includegraphics[width=0.24\textwidth]{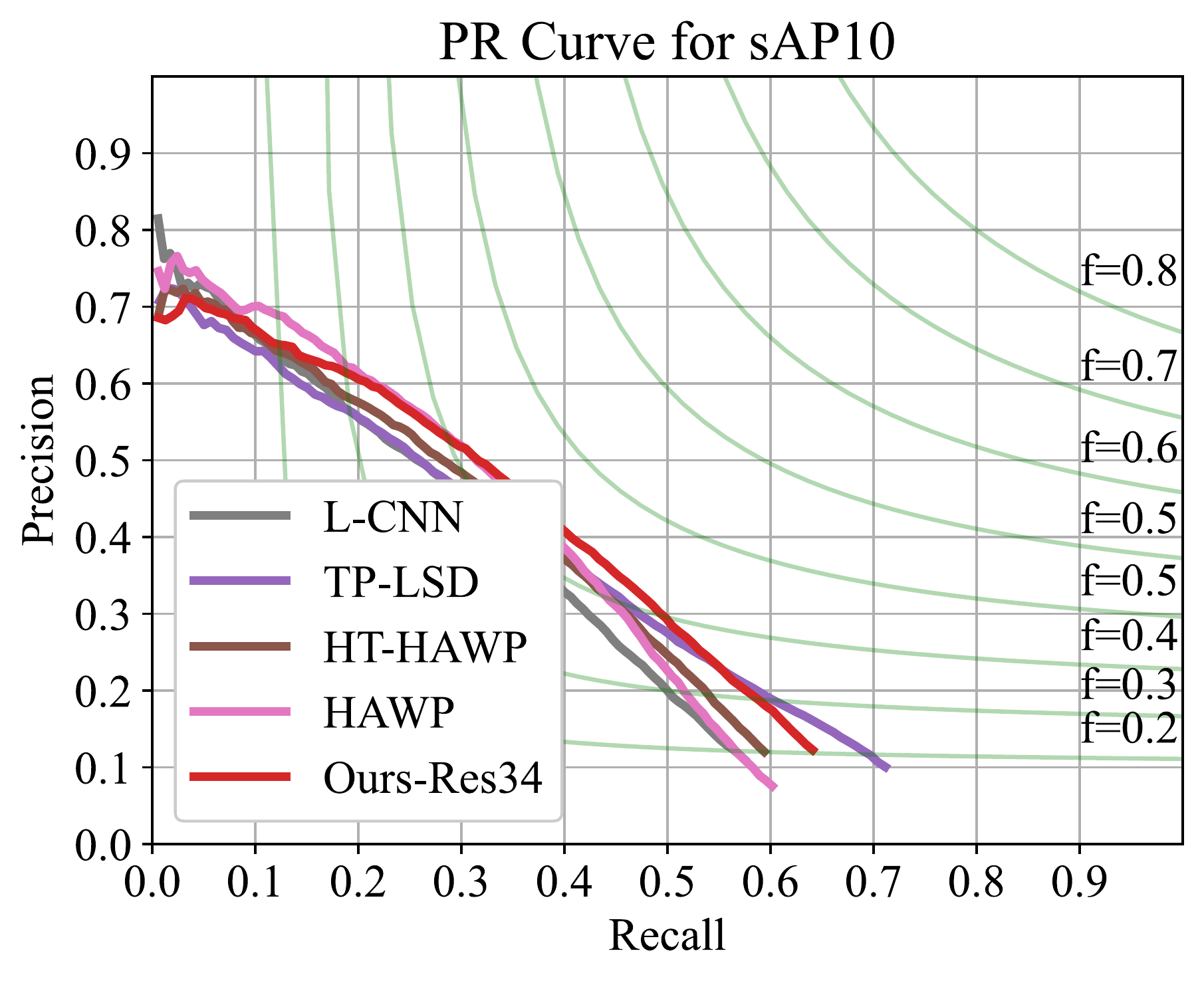}
\includegraphics[width=0.24\textwidth]{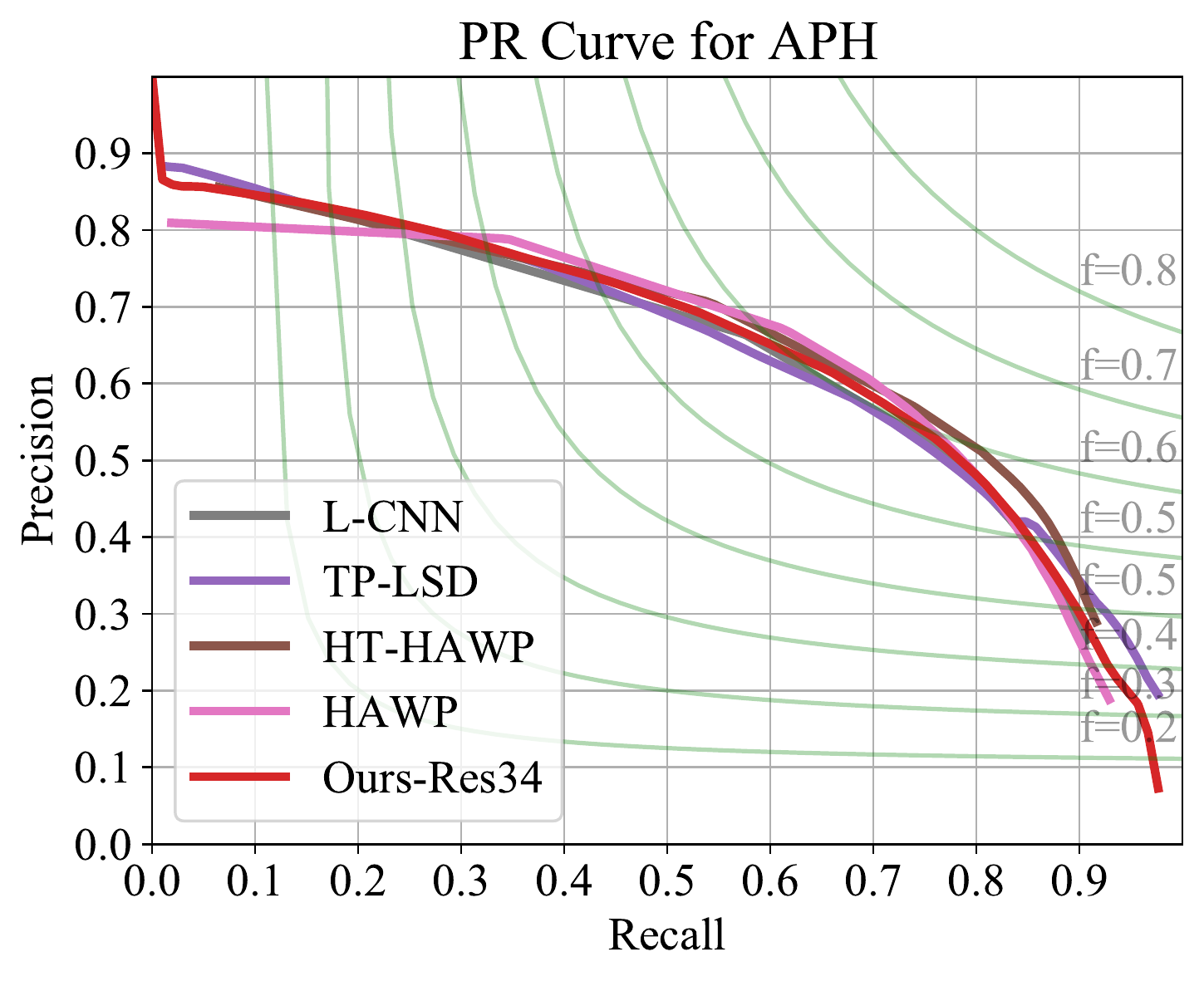}
\caption{PR curves of sAP$^{10}$ and AP$^H$ on Wireframe datasets (the left two figures) and YorkUrban datasets (the right two figures). The curve of our model is depicted in red. The results of DWP, AFM, and LSD on YorkUrban datasets are not displayed since they are slightly lower than the current methods.}
    \label{PR}
\end{figure*}

\subsubsection{Line Descriptor Loss}
We utilize the triplet loss proposed in Facenet\cite{Triplet} to learn a line descriptor. Since the descriptors are regularized by $l_2$ normalization, the cosine similarity of two descriptors can be represented as $cos(d_i,d_j) = d^T_id_j$, where $d_i, d_j$ are two descriptors. Given image pair $(I^A, I^B)$ and their line segments set $L^A, L^B$, let $L^A_i, d^A_i$ be the $i$-th line segment of image $I^A$ and its corresponding descriptor, $d_i^+$ be the descriptor of its matched line segment in image $I^B$, $d_i^-$ be the descriptor of its unmatched line segment in image $I^B$ with the maximal cosine similarity. Then the hard-negative triplet loss from image $I^A$ to $I^B$ can be represented as:
\begin{small}
\begin{equation}
\mathcal{T}(A,B)=\frac{1}{N}\sum_{i=1}^N[m-d_i^Td_i^++d_i^Td_i^-]_+
\label{triplet_loss}
\end{equation}
\end{small}
where $[x]_{+}=\max \{0, x\}$. $N$ is the number of line segments in $A$, $m$ is the margin that simultaneously enhances the  consistency of matched line segments and the discrepancy of unmatched line segments.
As mentioned in Section \ref{static_and_dynamic}, we have both static and dynamic line segments, so the overall loss of descriptor loss is:
\begin{small}
\begin{equation}
\begin{split}
    \mathcal{L}_{d}(A,B)=&\lambda_D(\mathcal{T}^D(A,B)+\mathcal{T}^D(B,A))+ \\  &\lambda_S(\mathcal{T}^S(A,B)+\mathcal{T}^S(B,A))
\end{split}
\end{equation}
\end{small}
where $\mathcal{T}^D, \mathcal{T}^S$ represent the dynamic and static descriptor loss according to Eq \eqref{triplet_loss}. We set $\lambda_D = \sqrt{\frac{e}{E}}$ and $\lambda_S = 1-\sqrt{\frac{e}{E}}$ in this paper, where the $E$ denotes the total epochs of entire training process and $e$ denotes the current epoch. Briefly, we expect to rely more on static loss at the early training stage, and rely more on dynamic loss after the detector is well trained to adapt the descriptor to the actual detection result.
\section{Experiments}

\subsection{Experiment Setting}

\textbf{Implementation details:}
We use ResNet34\cite{ResNet} and optionally Hourglass Network\cite{Hourglass} and as the backbone, respectively. We conduct standard data augmentation for the training set, including horizontal/vertical flip and random rotate. Input images are resize to $ 512 \times 512$.  Our model is trained using ADAM\cite{Adam} optimizer with a total of 170 epochs on four NVIDIA RTX 2080Ti GPUs and an Inter Xeon Gold 6130 2.10 GHz CPU. The initial learning rate, weight decay, and batch size are set to $ 1e-3 $, $ 1e-5 $, and 16 respectively. The learning rate is divided by 10 at the 100th and 150th epoch.

\textbf{Datasets:}
We train and evaluate our model on Wireframe Dataset\cite{wireframe_cvpr18}, which contains 5000 images for training and 462 images for testing. We further evaluate on YorkUrban dataset\cite{York} with 102 test images from both indoor scenes and outdoor scenes to validate the generalization ability.

\textbf{Structural Average Precision Metric\cite{LCNN}:} The structural average precision (sAP) of the line segment is based on the L2-distance between the predicted end-points and the ground truths. The predicted line segments will be counted as 
True Positive (TP) if the distance is less than a certain threshold $\vartheta$ and otherwise False Positives (FP). We set the threshold $ \vartheta = 5,10,15 $ and report the corresponding results, denote by sAP$ ^5$, sAP$^{10}$, sAP$^{15}$. For more details see \cite{LCNN}.

\textbf{Heatmap based Metric\cite{LCNN}:} Heatmap-based F-score and average precision, $ F^H $ and $ AP^H $ are typical metrics used in wireframe parsing and line segment detection. We first convert the predicted line and ground truth line to two heatmaps by rasterizing the lines respectively. Then we can calculate the pixel-level precision and recall (PR) curves. Finally, we can compute $F^H $ and $ AP^H $ with the PR curves.

\subsection{Comparison Experiments on Line Detection}

We compare our proposed ELSD with line segment detection methods and wireframe parsing methods. Our model use ResNet34 as backbone and for a fair comparison with other methods, we also alter the backbone with Hourglass denote by Ours-HG. Ours-Lite is a faster version of our model. In Ours-Lite, we resize the input image to $ 256 \times 256 $ and add a decoder in backbone. Therefore the outputs maps of each head is $ 256 \times 256 $. Table \ref{CompareWithLSD} shows quantitative results based on sAP, $AP^H$, $F^H$, and FPS of line segment detection. 

Ours-Res34 model achieves the best sAP on two datasets at a FPS of 42.6. It outperforms HAWP by 2.3\% and 1.8\% in msAP(mean of sAP) metric on Wireframe and YorkUrban respectively. Besides, when we replace the backbone with an Hourglass network(Ours-HG), it stills reaches a comparable sAP results on Wireframe. Since the HAWP and L-CNN are two stage methods, their inference speeds are limited. Moreover, their line segments rely on a pair of junctions, where junctions are usually local features that contain less global information. On the other hand, benefiting from more accurate midpoint detection and a more compact line representation method, our method is superior to TP-LSD. For further comparison, we evaluate the AP of mid-points similar to Junction AP proposed in L-CNN\cite{LCNN}. The mean AP of mid-point of ELSD is 2.9\% higher than TP-LSD, which means the mid-points are predicted more accurately in ELSD. 
 
In terms of the heatmap based metric, ELSD shows significant results in AP$^H$=87.2 on wireframe dataset and  achieves comparable results on F$^H$. Since our model predicts the line's angle, the angle prediction error of only one line segment could produce a lot of incorrect pixels, and but has the less influence to sAP. Therefore, the improvement of our model in pixel-based metric is not as obvious as that of sAP.

Our lightweight model can reach 107.5 FPS, which is 1.4 $\sim$ 48.9 times faster than other learning-based methods while the accuracy drop is limited. We use Ours-Res34 as the representative model and depicted the precision and recall curves on both datasets in Figure \ref{PR}. Our ELSD outperforms other line segment detection methods especially in sAP metric on Wireframe dataset. Besides, ELSD achieved better generalization ability on YorkUrban dataset than other two-stage methods.

\subsection{Ablation Study for Line Detection}\label{ablation_sec}
We run ablation experiments on the Wireframe dataset, as reported in Table \ref{ablation study}.

\textbf{NCS:} NCS is to suppress the midpoints of fragmented line segments and remain the midpoints of entire segments. It improves the $sAP^{10}$ from 0.680 to 0.689 according to No.3 and No.1 . 

\textbf{Descriptor:} The multi-task learning of detection and description leads to very small reduction on the detection accuracy $sAP^{10}$ from 0.689 to 0.685, according to No.1 and No.2 . 

\textbf{Upsample:} For detecting line segments in real time, we use the shared feature map with 128 resolution, which is the same setting as L-CNN and HAWP. However, the prediction of the center in 128 resolution is much difficult than higher resolution. We solve this problem by upsampling the midpoint map, centerness map, geometrics maps and fine offset maps to 256 resolution. The $sAP^{10}$ is thus improved from 0.658 to 0.689 according to No.4 and No.1. Since we only upsample once by bilinear interpolation or deconvolution, it has almost no extra cost on inference speed.

\textbf{Focal loss:} We use a variant focal loss instead of standard Binary Cross Entropy (BCE) loss for training the midpoint map. Since we treat the prediction of the midpoints as a binary classification problem, the focal loss that we used can have the ability to focus on the hard classified examples of midpoints. By introducing the focal loss, the $ sAP^{10}$ is improved from 0.660 to 0.689 according to No.5 and No.1 . 

\textbf{CAL:} The proposed CAL representation is compared to the Tri-points representation in TP-LSD. The $sAP^{10}$ is improved from 0.679 to 0.689 according to No.6 and No.1 by replacing Tri-points with the CAL representation. This is because the Tri-points need to regress more parameters than the CAL representation (4 vs 2), and the angle is easier to learn than the displacements.

\begin{table}[]
\begin{center}
\scalebox{0.65}{
\begin{tabular}{|c|c|c|c|c|c|c|c|c|}
\hline
No. & NCS & Upsample & Focal loss & CAL & Descriptor & sAP$^5$          & sAP$^{10}$         & sAP$^{15}$         \\ \hline
1   & \checkmark   & \checkmark        & \checkmark          & \checkmark     &            & \textbf{64.3} & \textbf{68.9} & \textbf{70.9} \\ \hline
2   & \checkmark   & \checkmark        & \checkmark          & \checkmark     & \checkmark          & 64.2          & 68.5          & 70.3          \\ \hline
3   &     & \checkmark        & \checkmark          & \checkmark     &            & 63.6          & 68.0          & 70.0          \\ \hline
4   & \checkmark   &          & \checkmark          & \checkmark     &            & 60.3          & 65.8          & 68.2          \\ \hline
5   & \checkmark   & \checkmark        &            & \checkmark     &            & 61.8          & 66.0          & 68.0          \\ \hline
6   & \checkmark   & \checkmark        & \checkmark          &       &            & 62.3          & 67.9          & 70.3          \\ \hline
7   &     &          &            &       &            & 58.0          & 62.8          & 64.8          \\ \hline
\end{tabular}
}
\end{center}
\caption{Ablation study of ELSD. See text for details.}
\label{ablation study}
\end{table}

\subsection{Comparison Experiments on Line Description}
To evaluate the line descriptor performance, we compare our method with LBD\cite{LBD} and LLD\cite{LLD}. The methods \cite{LJL,GNN} etc, are not involved to be compared, because they leverage the additional geometric characters of lines, other than local appearances. We test all of the algorithms on a subset of ScanNet dataset\cite{scannet} which is an RGB-D video dataset annotated with 3D camera poses. We select about 1000 image pairs with large viewpoint change, rotation change, and scale change for quantitative evaluation. We further compute the corresponding line descriptors of line segments detected by our model. We then obtain the ground truth line matches of the image pairs by checking if the reprojection error of corresponding lines less than a certain threshold.
We find the nearest neighbors to match across descriptors and perform cross-checking then we can get predicted corresponds of line segments.
We report the recall, precision and F-score to evaluate different descriptors. In our experiments, we use the OpenCV implementation of 72-dimensional LBD descriptors and the pre-trained model of LLD descriptors provided by the author. Meanwhile, our model is trained with setting the length of the descriptor to 256, 64 and 36 respectively.

The results are shown in Table \ref{desc_table}. Our descriptors outperform the LBD and LLD significantly, especially in Recall. 
The LBD descriptor is designed by the human priority that might not be the optimal solution. The LLD descriptor and the similar learning-based descriptors\cite{WLD,DLD} are trained with the line segments given by the line detectors such as Edlines\cite{edline}. However, there is a gap between those detected line segments and the annotated line segments in the datasets\cite{York,wireframe_cvpr18}.
In comparison, our descriptors cooperate well with our line detector since they share most of the parameters and representation, and their training is coupled, which can further reduce computation cost. The overall inference speed of ELSD (ResNet34 as backbone) with both line detector and line descriptor can achieve 38 FPS. 
Moreover, the 64-dimensional descriptor presents the same result as the 256-dimensional descriptor, and is better than the 36-dimensional descriptor in accuracy.
\begin{table}[]
\begin{center}
\scalebox{0.8}{
\begin{tabular}{lcccc}
\hline
\multicolumn{1}{c}{Methods} & Dimension & Precision(\%) & Recall(\%) & F-Score(\%) \\ \hline
LBD                         & 78        & 69.3          & 63.8       & 66.4        \\ \hline
LLD                         & 64        & 57.5          & 43.6       & 49.6        \\ \hline
WLD                         & 64        & 57.5          & 43.6       & 49.6        \\ \hline
Ours                        & 256       & 72.6          & \textbf{77.1}       & 74.7        \\
                            & 64        & \textbf{73.5}          & 76.2       & \textbf{74.8}        \\
                            & 36        & 72.2          & 75.3       & 73.7        \\ \hline
\end{tabular}
}
\end{center}
\caption{The precision, recall and F-Score for LBD, LLD, and Ours with different dimension.}
\label{desc_table}
\end{table}

\section{Conclusion}
This paper proposes a fast and accurate model ELSD that simultaneously detects line segments and their descriptors in a single forward pass, allowing share computation and representation in the two tasks. To detect line segments, We first utilize the Center-Angle-Length (CAL) representation to encode a line segment that fully exploits the geometric characters of lines. Furthermore, a centerness map is introduced to filter the false line segments by Non-Centerness-Suppression (NCS). Our proposed line detector achieves state-of-the-art performance on two benchmarks in both accuracy and efficiency. Moreover, our model also achieves real-time speed with a single GPU. The lite model can reach the high speed of 107.5 FPS while keeping a comparable performance, and thus is useful for many higher-level tasks such as SLAM and SfM that require high real-time performance.

\newpage
{\small
\bibliographystyle{ieee_fullname}
\bibliography{egbib}

\begin{thebibliography}{10}\itemsep=-1pt

\bibitem{edline}
C. {Akinlar} and C. {Topal}.
\newblock Edlines: Real-time line segment detection by edge drawing (ed).
\newblock In {\em 2011 18th IEEE International Conference on Image Processing},
  pages 2837--2840, 2011.

\bibitem{scannet}
Angela Dai, Angel~X. Chang, Manolis Savva, Maciej Halber, Thomas Funkhouser,
  and Matthias Nie{\ss}ner.
\newblock Scannet: Richly-annotated 3d reconstructions of indoor scenes.
\newblock In {\em Proc. Computer Vision and Pattern Recognition (CVPR), IEEE},
  2017.

\bibitem{York}
Patrick Denis, James~H. Elder, and Francisco~J. Estrada.
\newblock Efficient edge-based methods for estimating manhattan frames in urban
  imagery.
\newblock In David Forsyth, Philip Torr, and Andrew Zisserman, editors, {\em
  Computer Vision -- ECCV 2008}, pages 197--210, Berlin, Heidelberg, 2008.
  Springer Berlin Heidelberg.

\bibitem{superpoint}
Daniel DeTone, Tomasz Malisiewicz, and Andrew Rabinovich.
\newblock Superpoint: Self-supervised interest point detection and description.
\newblock In {\em Proceedings of the IEEE conference on computer vision and
  pattern recognition workshops}, pages 224--236, 2018.

\bibitem{LSD}
Rafael Gioi, Jeremie Jakubowicz, Jean-Michel Morel, and Gregory Randall.
\newblock Lsd: A fast line segment detector with a false detection control.
\newblock {\em IEEE transactions on pattern analysis and machine intelligence},
  32:722--32, 04 2010.

\bibitem{roipool}
Ross Girshick.
\newblock Fast r-cnn.
\newblock In {\em Proceedings of the IEEE international conference on computer
  vision}, pages 1440--1448, 2015.

\bibitem{roialign}
Kaiming He, Georgia Gkioxari, Piotr Doll{\'a}r, and Ross Girshick.
\newblock Mask r-cnn.
\newblock In {\em Proceedings of the IEEE international conference on computer
  vision}, pages 2961--2969, 2017.

\bibitem{ResNet}
Kaiming He, Xiangyu Zhang, Shaoqing Ren, and Jian Sun.
\newblock Deep residual learning for image recognition.
\newblock In {\em Proceedings of the IEEE Conference on Computer Vision and
  Pattern Recognition (CVPR)}, June 2016.

\bibitem{plvio}
Yijia He, Ji Zhao, Yue Guo, Wenhao He, and Kui Yuan.
\newblock Pl-vio: Tightly-coupled monocular visual--inertial odometry using
  point and line features.
\newblock {\em Sensors}, 18(4):1159, 2018.

\bibitem{planar_SLAM}
Ming Hsiao, Eric Westman, Guofeng Zhang, and Michael Kaess.
\newblock Keyframe-based dense planar slam.
\newblock In {\em 2017 IEEE International Conference on Robotics and Automation
  (ICRA)}, pages 5110--5117. IEEE, 2017.

\bibitem{wireframe_cvpr18}
Kun Huang, Yifan Wang, Zihan Zhou, Tianjiao Ding, Shenghua Gao, and Yi Ma.
\newblock Learning to parse wireframes in images of man-made environments.
\newblock In {\em CVPR}, June 2018.

\bibitem{TP-LSD}
Siyu Huang, Fangbo Qin, Pengfei Xiong, Ning Ding, Yijia He, and Xiao Liu.
\newblock {TP-LSD:} tri-points based line segment detector.
\newblock In {\em Computer Vision - {ECCV} 2020 - 16th European Conference,
  Glasgow, UK, August 23-28, 2020, Proceedings, Part {XXVII}}, Lecture Notes in
  Computer Science. Springer, 2020.

\bibitem{RGB-D_SLAM_for_planar}
Pyojin Kim, Brian Coltin, and H~Jin Kim.
\newblock Linear rgb-d slam for planar environments.
\newblock In {\em Proceedings of the European Conference on Computer Vision
  (ECCV)}, pages 333--348, 2018.

\bibitem{Adam}
Diederik~P. Kingma and Jimmy Ba.
\newblock Adam: A method for stochastic optimization, 2017.

\bibitem{WLD}
M. Lange, Claudio Raisch, and A. Schilling.
\newblock Wld: A wavelet and learning based line descriptor for line feature
  matching.
\newblock In {\em VMV}, 2020.

\bibitem{DLD}
M. {Lange}, F. {Schweinfurth}, and A. {Schilling}.
\newblock Dld: A deep learning based line descriptor for line feature matching.
\newblock In {\em 2019 IEEE/RSJ International Conference on Intelligent Robots
  and Systems (IROS)}, pages 5910--5915, 2019.

\bibitem{CornerNet}
Hei Law and Jia Deng.
\newblock Cornernet: Detecting objects as paired keypoints.
\newblock In {\em Proceedings of the European Conference on Computer Vision
  (ECCV)}, September 2018.

\bibitem{LJL}
Kai Li, Jian Yao, Xiaohu Lu, Li Li, and Zhichao Zhang.
\newblock Hierarchical line matching based on line-junction-line structure
  descriptor and local homography estimation.
\newblock {\em Neurocomputing}, 184, 01 2016.

\bibitem{Focal}
Tsung-Yi Lin, Priya Goyal, Ross Girshick, Kaiming He, and Piotr Dollar.
\newblock Focal loss for dense object detection.
\newblock In {\em Proceedings of the IEEE International Conference on Computer
  Vision (ICCV)}, Oct 2017.

\bibitem{HT-HAWP}
Yancong Lin, Silvia~L. Pintea, and Jan~C. van Gemert.
\newblock Deep hough-transform line priors.
\newblock In Andrea Vedaldi, Horst Bischof, Thomas Brox, and Jan-Michael Frahm,
  editors, {\em Computer Vision -- ECCV 2020}, 2020.

\bibitem{SIFT}
David~G Lowe.
\newblock Object recognition from local scale-invariant features.
\newblock In {\em Proceedings of the seventh IEEE international conference on
  computer vision}, volume~2, pages 1150--1157. Ieee, 1999.

\bibitem{GNN}
QuanMeng Ma, Guang Jiang, and DianZhi Lai.
\newblock Robust line segments matching via graph convolution networks, 2020.

\bibitem{Hourglass}
Alejandro Newell, Kaiyu Yang, and Jia Deng.
\newblock Stacked hourglass networks for human pose estimation.
\newblock In {\em ECCV (8)}, pages 483--499, 2016.

\bibitem{orb}
Ethan Rublee, Vincent Rabaud, Kurt Konolige, and Gary Bradski.
\newblock Orb: An efficient alternative to sift or surf.
\newblock In {\em 2011 International conference on computer vision}, pages
  2564--2571. Ieee, 2011.

\bibitem{Triplet}
Florian Schroff, Dmitry Kalenichenko, and James Philbin.
\newblock Facenet: A unified embedding for face recognition and clustering.
\newblock In {\em Proceedings of the IEEE Conference on Computer Vision and
  Pattern Recognition (CVPR)}, June 2015.

\bibitem{FCOS}
Zhi Tian, Chunhua Shen, Hao Chen, and Tong He.
\newblock {FCOS}: Fully convolutional one-stage object detection.
\newblock In {\em Proc. Int. Conf. Computer Vision (ICCV)}, 2019.

\bibitem{LLD}
A. {Vakhitov} and V. {Lempitsky}.
\newblock Learnable line segment descriptor for visual slam.
\newblock {\em IEEE Access}, 7:39923--39934, 2019.

\bibitem{MSLD}
Zhiheng Wang, Fuchao Wu, and Zhanyi Hu.
\newblock Msld: A robust descriptor for line matching.
\newblock {\em Pattern Recognition}, 42(5):941--953, 2009.

\bibitem{polarnet}
Wu Xiongwei, Steven HOI, and Doyen Sahoo.
\newblock Polarnet: Learning to optimize polar keypoints for keypoint based
  object detection.
\newblock In {\em International Conference on Learning Representations}, 2021.

\bibitem{Line_Correspondences}
C. {Xu}, L. {Zhang}, L. {Cheng}, and R. {Koch}.
\newblock Pose estimation from line correspondences: A complete analysis and a
  series of solutions.
\newblock {\em IEEE Transactions on Pattern Analysis and Machine Intelligence},
  39(6):1209--1222, 2017.

\bibitem{LETR}
Yifan Xu, Weijian Xu, David Cheung, and Zhuowen Tu.
\newblock Line segment detection using transformers without edges, 2021.

\bibitem{AFM}
Nan Xue, Song Bai, Fudong Wang, Gui-Song Xia, Tianfu Wu, and Liangpei Zhang.
\newblock Learning attraction field representation for robust line segment
  detection.
\newblock In {\em Proceedings of the IEEE/CVF Conference on Computer Vision and
  Pattern Recognition (CVPR)}, June 2019.

\bibitem{HAWP}
Nan Xue, Tianfu Wu, Song Bai, Fudong Wang, Gui-Song Xia, Liangpei Zhang, and
  Philip~H.S. Torr.
\newblock Holistically-attracted wireframe parsing.
\newblock In {\em Proceedings of the IEEE/CVF Conference on Computer Vision and
  Pattern Recognition (CVPR)}, June 2020.

\bibitem{Line-Based_Map}
G. {Zhang}, J.~H. {Lee}, J. {Lim}, and I.~H. {Suh}.
\newblock Building a 3-d line-based map using stereo slam.
\newblock {\em IEEE Transactions on Robotics}, 31(6):1364--1377, 2015.

\bibitem{LBD}
Lilian Zhang and Reinhard Koch.
\newblock An efficient and robust line segment matching approach based on lbd
  descriptor and pairwise geometric consistency.
\newblock {\em Journal of Visual Communication and Image Representation},
  24(7):794--805, 2013.

\bibitem{PPGNET}
Ziheng Zhang, Zhengxin Li, Ning Bi, Jia Zheng, Jinlei Wang, Kun Huang, Weixin
  Luo, Yanyu Xu, and Shenghua Gao.
\newblock Ppgnet: Learning point-pair graph for line segment detection.
\newblock In {\em Proceedings of the IEEE/CVF Conference on Computer Vision and
  Pattern Recognition (CVPR)}, June 2019.

\bibitem{Centernet}
Xingyi Zhou, Dequan Wang, and Philipp Krähenbühl.
\newblock Objects as points, 2019.

\bibitem{LCNN}
Yichao Zhou, Haozhi Qi, and Yi Ma.
\newblock End-to-end wireframe parsing.
\newblock In {\em Proceedings of the IEEE/CVF International Conference on
  Computer Vision (ICCV)}, October 2019.

\bibitem{3Dwireframe}
Yichao Zhou, Haozhi Qi, Yuexiang Zhai, Qi Sun, Zhili Chen, Li-Yi Wei, and Yi
  Ma.
\newblock Learning to reconstruct 3d manhattan wireframes from a single image.
\newblock In {\em Proceedings of the IEEE/CVF International Conference on
  Computer Vision (ICCV)}, October 2019.

\end{thebibliography}


\begin{thebibliography}{10}\itemsep=-1pt

\bibitem{scannet}
Angela Dai, Angel~X. Chang, Manolis Savva, Maciej Halber, Thomas Funkhouser,
  and Matthias Nie{\ss}ner.
\newblock Scannet: Richly-annotated 3d reconstructions of indoor scenes.
\newblock In {\em Proc. Computer Vision and Pattern Recognition (CVPR), IEEE},
  2017.

\bibitem{York}
Patrick Denis, James~H. Elder, and Francisco~J. Estrada.
\newblock Efficient edge-based methods for estimating manhattan frames in urban
  imagery.
\newblock In David Forsyth, Philip Torr, and Andrew Zisserman, editors, {\em
  Computer Vision -- ECCV 2008}, pages 197--210, Berlin, Heidelberg, 2008.
  Springer Berlin Heidelberg.

\bibitem{LSD}
Rafael Gioi, Jeremie Jakubowicz, Jean-Michel Morel, and Gregory Randall.
\newblock Lsd: A fast line segment detector with a false detection control.
\newblock {\em IEEE transactions on pattern analysis and machine intelligence},
  32:722--32, 04 2010.

\bibitem{ResNet}
Kaiming He, Xiangyu Zhang, Shaoqing Ren, and Jian Sun.
\newblock Deep residual learning for image recognition.
\newblock In {\em Proceedings of the IEEE Conference on Computer Vision and
  Pattern Recognition (CVPR)}, June 2016.

\bibitem{wireframe_cvpr18}
Kun Huang, Yifan Wang, Zihan Zhou, Tianjiao Ding, Shenghua Gao, and Yi Ma.
\newblock Learning to parse wireframes in images of man-made environments.
\newblock In {\em CVPR}, June 2018.

\bibitem{Adam}
Diederik~P. Kingma and Jimmy Ba.
\newblock Adam: A method for stochastic optimization, 2017.

\bibitem{HT-HAWP}
Yancong Lin, Silvia~L. Pintea, and Jan~C. van Gemert.
\newblock Deep hough-transform line priors.
\newblock In Andrea Vedaldi, Horst Bischof, Thomas Brox, and Jan-Michael Frahm,
  editors, {\em Computer Vision -- ECCV 2020}, 2020.

\bibitem{Hourglass}
Alejandro Newell, Kaiyu Yang, and Jia Deng.
\newblock Stacked hourglass networks for human pose estimation.
\newblock In {\em ECCV (8)}, pages 483--499, 2016.

\bibitem{LLD}
A. {Vakhitov} and V. {Lempitsky}.
\newblock Learnable line segment descriptor for visual slam.
\newblock {\em IEEE Access}, 7:39923--39934, 2019.

\bibitem{HAWP}
Nan Xue, Tianfu Wu, Song Bai, Fudong Wang, Gui-Song Xia, Liangpei Zhang, and
  Philip~H.S. Torr.
\newblock Holistically-attracted wireframe parsing.
\newblock In {\em Proceedings of the IEEE/CVF Conference on Computer Vision and
  Pattern Recognition (CVPR)}, June 2020.

\bibitem{LBD}
Lilian Zhang and Reinhard Koch.
\newblock An efficient and robust line segment matching approach based on lbd
  descriptor and pairwise geometric consistency.
\newblock {\em Journal of Visual Communication and Image Representation},
  24(7):794--805, 2013.

\bibitem{LCNN}
Yichao Zhou, Haozhi Qi, and Yi Ma.
\newblock End-to-end wireframe parsing.
\newblock In {\em Proceedings of the IEEE/CVF International Conference on
  Computer Vision (ICCV)}, October 2019.

\end{thebibliography}
}

\end{document}


\title{Supplementary Material}

\maketitle
\ificcvfinal\thispagestyle{empty}\fi


\section{Supplementary Material}
\subsection{Implementation details}
Our ELSD uses the backbone of U-shape network that adopts ResNet34\cite{ResNet} as the encoder and optionally Hourglass Network\cite{Hourglass} as the backbone. We conduct standard data augmentation for the training set, including horizontal/vertical flip and random rotate. Input images are resized to $512 \times 512$. Our model is trained using ADAM\cite{Adam} optimizer with a total of 170 epochs on four NVIDIA RTX 2080Ti GPUs and an Inter Xeon Gold 6130 2.10 GHz CPU. The initial learning rate, weight decay, and batch size are set to $1e-3$, $ 1e-5 $, and 16 respectively. The learning rate is divided by 10 at the 100th and 150th epoch. It is recommended to train line detector firstly and then jointly train with descriptors, since the line descriptor branch is easier to learn compared to the line detector branch. See the source code in the supplementary file for more details. 

\subsection{Qualitative Results on Line Segment Detector}
\begin{figure*}[]
\centering
\scalebox{1.1}{
\begin{subfigure}[b]{0.13\textwidth}
         \centering
         \includegraphics[width=\textwidth]{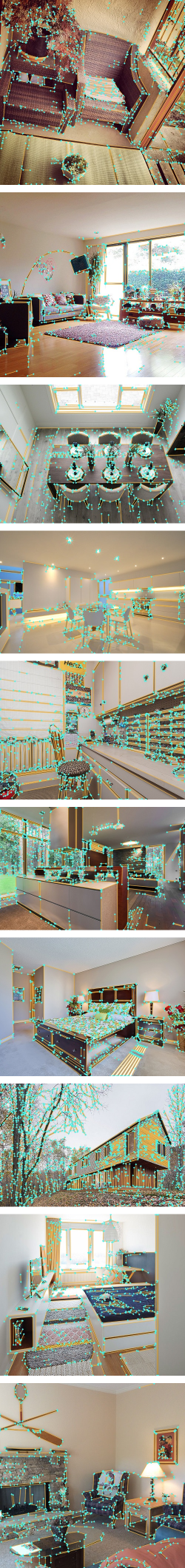}
         \caption{LSD}
\end{subfigure}
\begin{subfigure}[b]{0.13\textwidth}
         \centering
         \includegraphics[width=\textwidth]{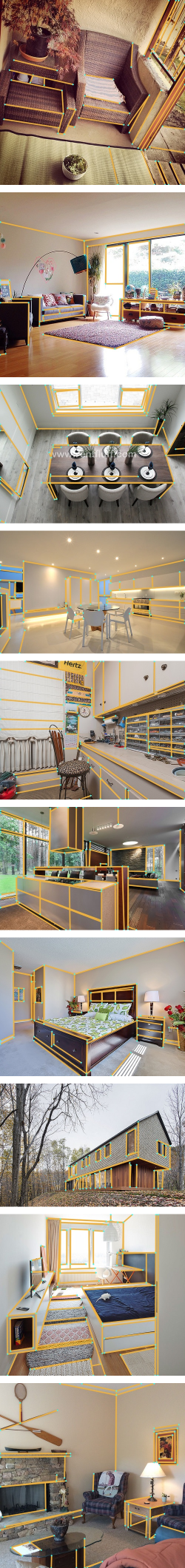}
         \caption{L-CNN}
\end{subfigure}
\begin{subfigure}[b]{0.13\textwidth}
         \centering
         \includegraphics[width=\textwidth]{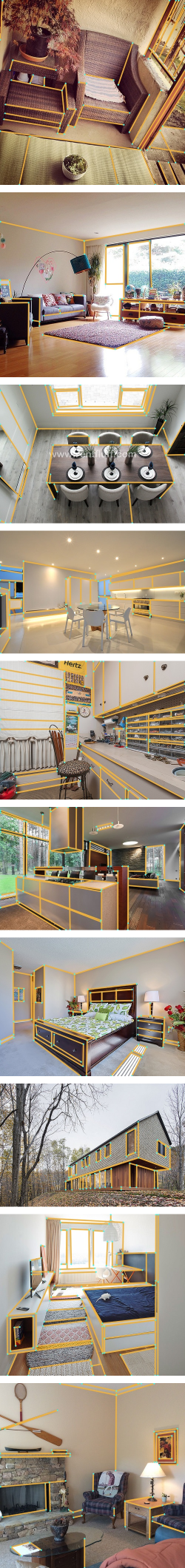}
         \caption{HAWP}
\end{subfigure}
\begin{subfigure}[b]{0.13\textwidth}
         \centering
         \includegraphics[width=\textwidth]{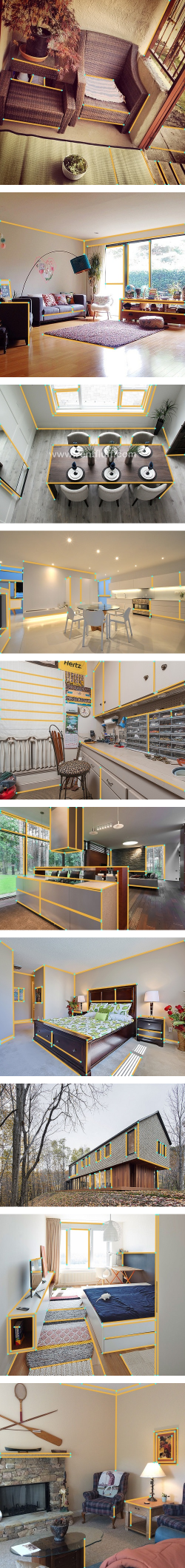}
         \caption{HT-HAWP}
\end{subfigure}
\begin{subfigure}[b]{0.13\textwidth}
         \centering
         \includegraphics[width=\textwidth]{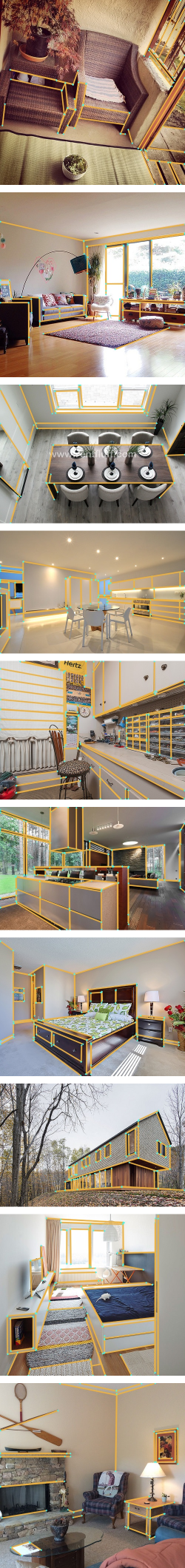}
         \caption{TP-LSD}
\end{subfigure}
\begin{subfigure}[b]{0.13\textwidth}
         \centering
         \includegraphics[width=\textwidth]{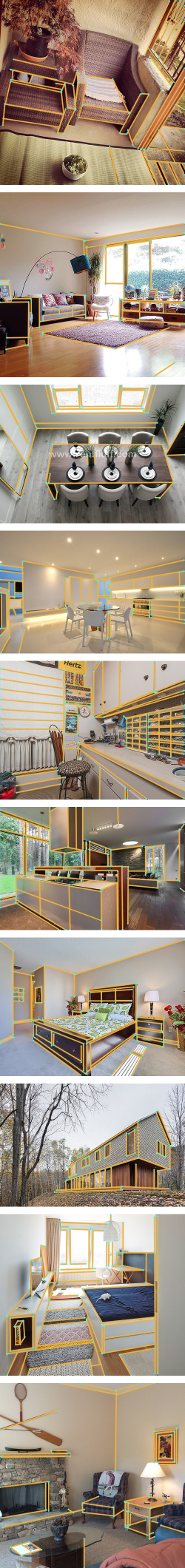}
         \caption{ELSD (Ours)}
\end{subfigure}
\begin{subfigure}[b]{0.13\textwidth}
         \centering
         \includegraphics[width=\textwidth]{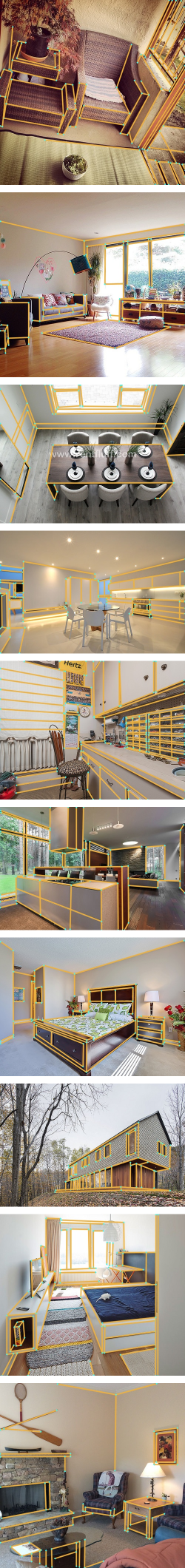}
         \caption{GT}
\end{subfigure}
}
\caption{Visualization of line detection methods on Wireframe dataset.}
\label{wire_vis}
\end{figure*}

\begin{figure*}[]
\centering
\scalebox{1.1}[1.2]{
\begin{subfigure}[b]{0.13\textwidth}
         \centering
         \includegraphics[width=\textwidth]{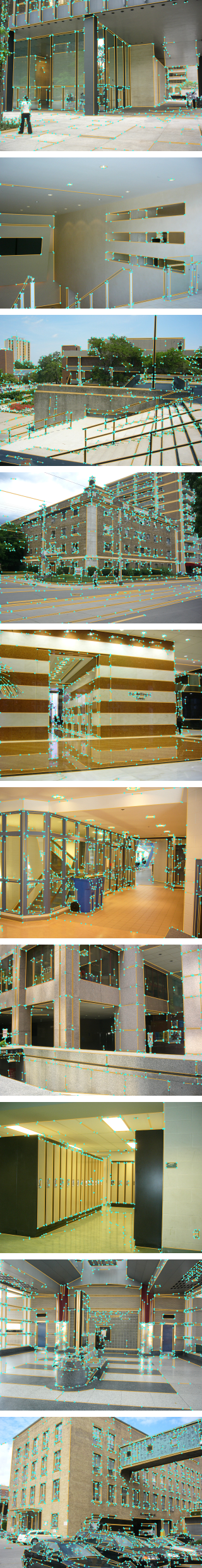}
         \caption{LSD}
\end{subfigure}
\begin{subfigure}[b]{0.13\textwidth}
         \centering
         \includegraphics[width=\textwidth]{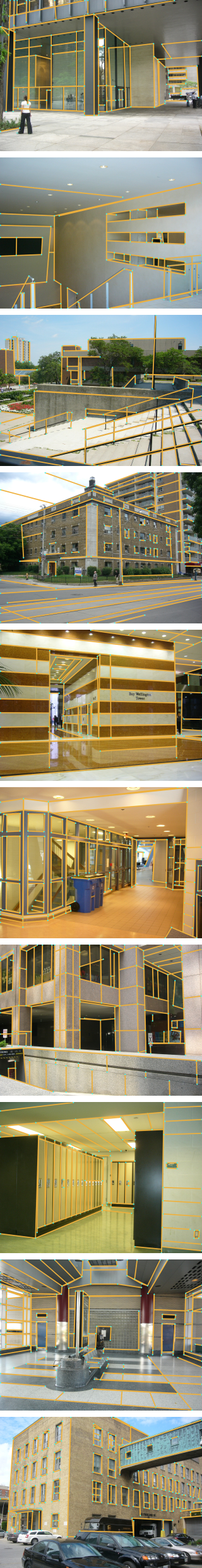}
         \caption{L-CNN}
\end{subfigure}
\begin{subfigure}[b]{0.13\textwidth}
         \centering
         \includegraphics[width=\textwidth]{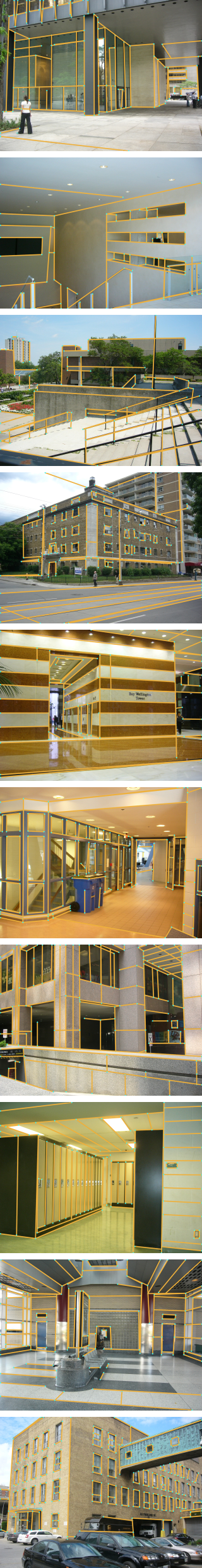}
         \caption{HAWP}
\end{subfigure}
\begin{subfigure}[b]{0.13\textwidth}
         \centering
         \includegraphics[width=\textwidth]{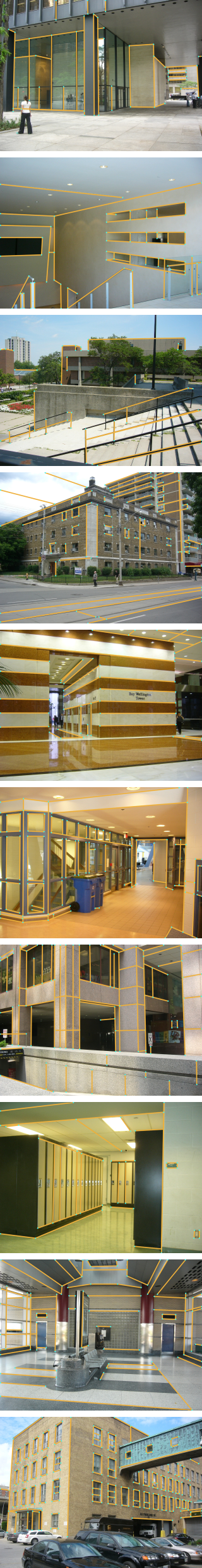}
         \caption{HT-HAWP}
\end{subfigure}
\begin{subfigure}[b]{0.13\textwidth}
         \centering
         \includegraphics[width=\textwidth]{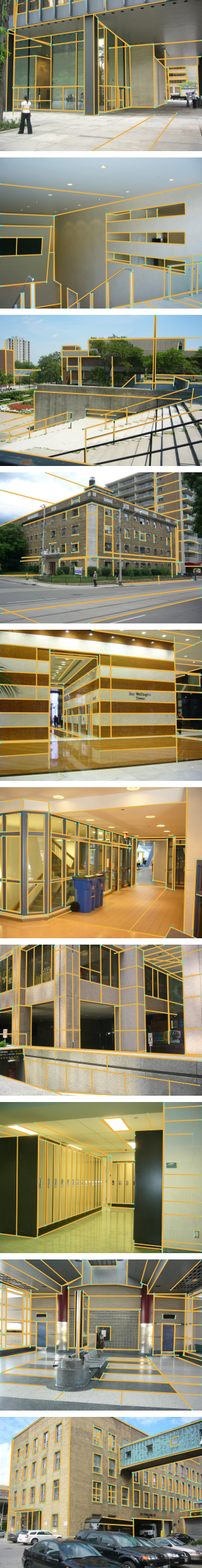}
         \caption{TP-LSD}
\end{subfigure}
\begin{subfigure}[b]{0.13\textwidth}
         \centering
         \includegraphics[width=\textwidth]{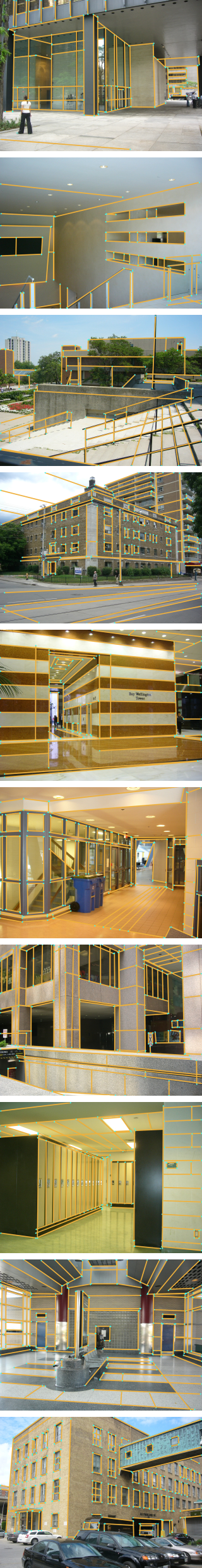}
         \caption{ELSD (Ours)}
\end{subfigure}
\begin{subfigure}[b]{0.13\textwidth}
         \centering
         \includegraphics[width=\textwidth]{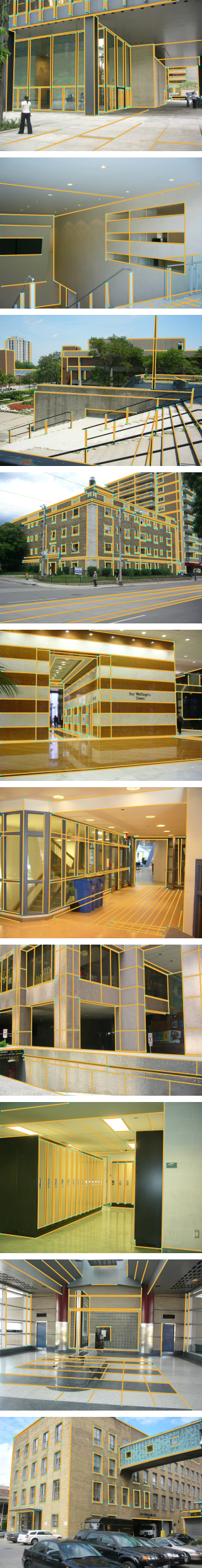}
         \caption{GT}
\end{subfigure}
}
\caption{Visualization of line detection methods on YorkUrban dataset.}
\label{york_vis}
\end{figure*}

We show more visualization results on the Wireframe dataset\cite{wireframe_cvpr18} and YorkUrban dataset\cite{York}
in Figure \ref{wire_vis} and Figure \ref{york_vis}. The configurations for visualization of different methods are as follows:
\begin{itemize}
\setlength{\itemsep}{0pt}
\setlength{\parsep}{0pt}
\setlength{\parskip}{0pt}
    \item The $a$-$contrario$ validation of LSD\cite{LSD} is set to $ -log\  \epsilon = 0.01 \times 1.75^{8}$.
    \item The thresholds in line verification of L-CNN\cite{LCNN}, HAWP\cite{HAWP} and HT-HAWP\cite{HT-HAWP} are set to 0.98, 0.95 and 0.99 respectively, where the PR curve of $sAP^{10}$ achieves maximum F-score on Wireframe dataset.
    \item The threshold of  root-point detection in TP-LSD is set as 0.43, where the PR curve of $sAP^{10}$ also achieves maximum F-score.
    \item For Our ELSD, the threshold of mid-point's score after Non-Centerness Suppression  is set to 0.22 for the same purpose.
    
\end{itemize}
\subsection{Qualitative Results on Line Descriptor}
\begin{figure*}[htbp]
\centering
\scalebox{1.0}[1.2]{
\begin{subfigure}[b]{0.24\textwidth}
         \centering
         \includegraphics[width=\textwidth]{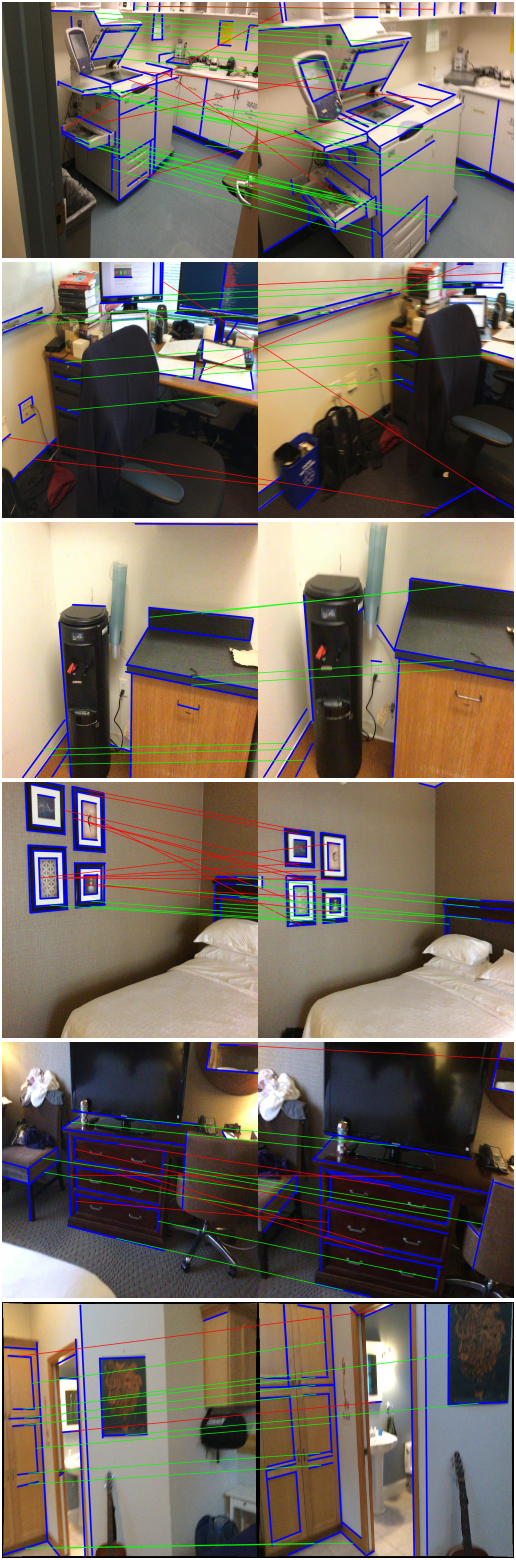}
         \caption{LBD}
\end{subfigure}
\begin{subfigure}[b]{0.24\textwidth}
         \centering
         \includegraphics[width=\textwidth]{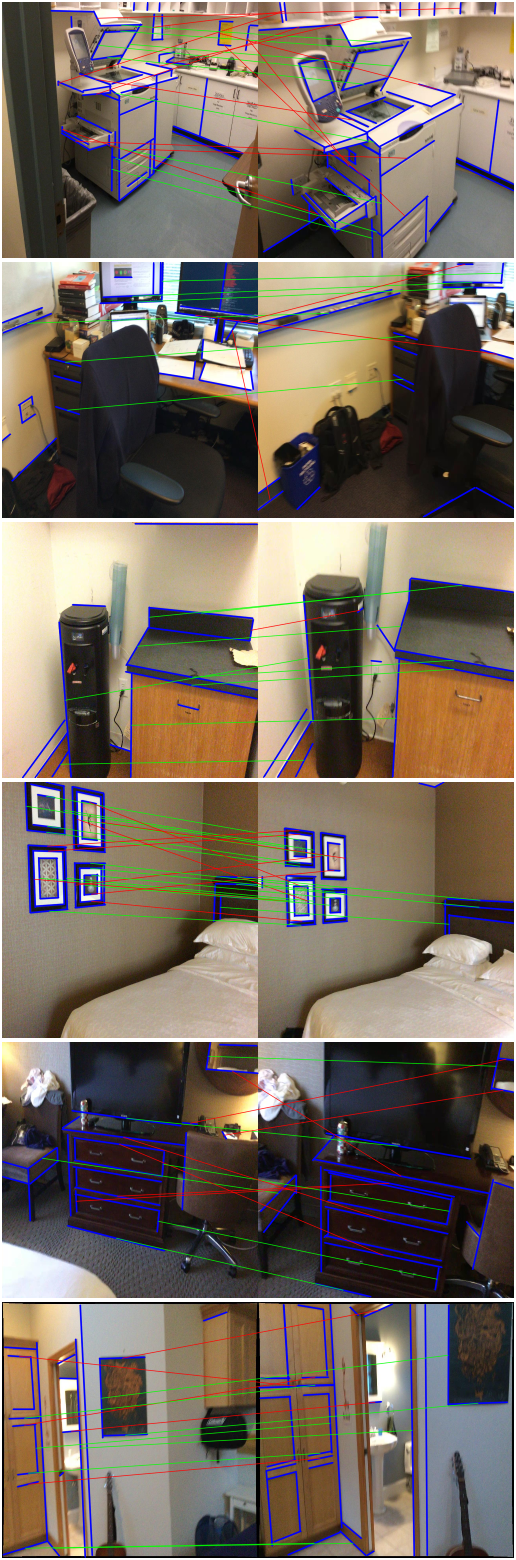}
         \caption{LLD}
\end{subfigure}
\begin{subfigure}[b]{0.24\textwidth}
         \centering
         \includegraphics[width=\textwidth]{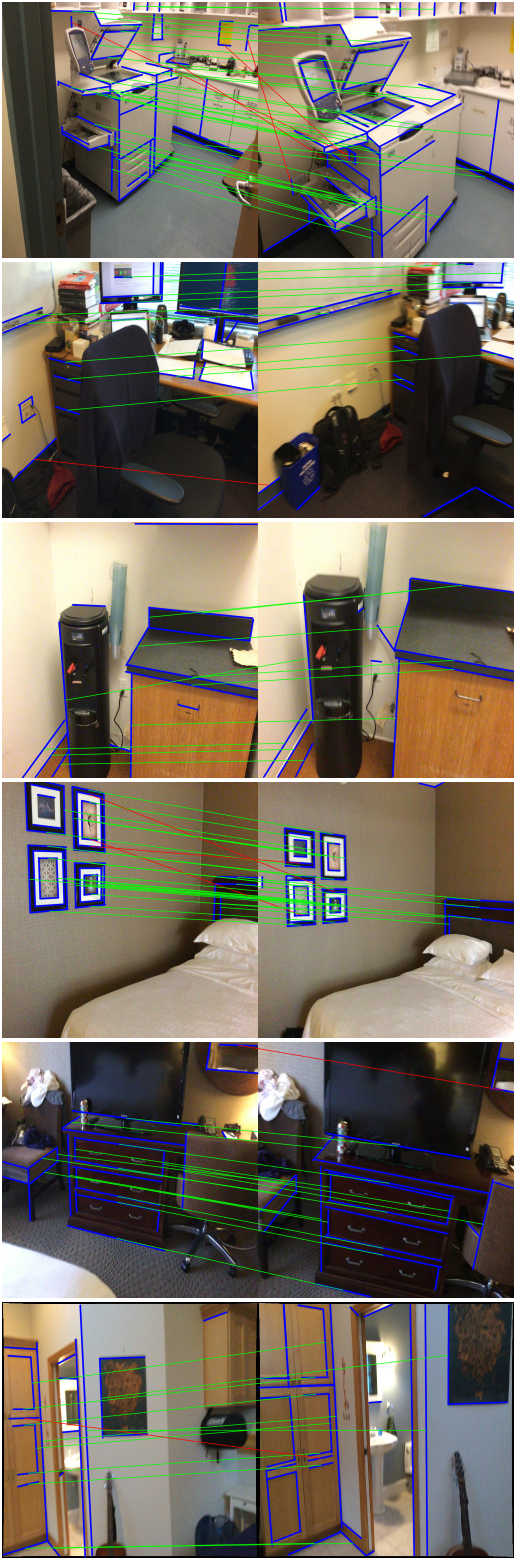}
         \caption{ELSD (Ours)}
\end{subfigure}
\begin{subfigure}[b]{0.24\textwidth}
         \centering
         \includegraphics[width=\textwidth]{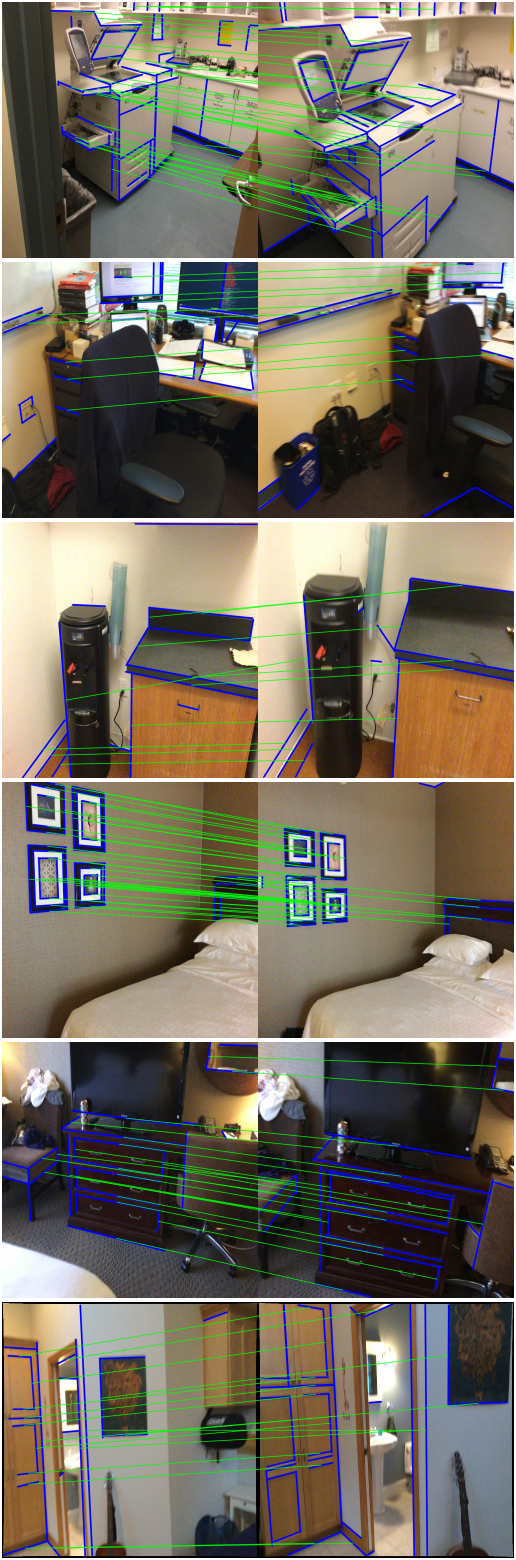}
         \caption{GT}
\end{subfigure}
}
\caption{Line matching results using descriptors. Note that the blue lines are predicted by our ELSD, and the green / red lines represent the true / false positive matches.}
\label{match_figure}
\end{figure*}

To perform the quantitative and qualitative evaluation for line matching using different descriptors, we select about 1000 image pairs from ScanNet\cite{scannet} dataset that includes large viewpoint change, rotation change, and scale change. We further visualize the line matching results of LBD\cite{LBD}, LLD\cite{LLD} and our 64-dimensional descriptor. We use the OpenCV implementation of 72-dimensional LBD descriptors and the official model of LLD descriptors. Note that we find the nearest neighbors to match lines across descriptors and perform cross-checking. The results are shown in Figure \ref{match_figure}.
{\small
\bibliographystyle{ieee_fullname}
\bibliography{egbib}
}

